\documentclass[journal]{IEEEtran}
\usepackage{amsmath,amsfonts}
\usepackage{algorithm}
\usepackage{array}
\usepackage{textcomp}
\usepackage{stfloats}
\usepackage{url}
\usepackage{verbatim}
\usepackage{graphicx}
\usepackage{cite}
\usepackage{orcidlink}
\usepackage{multirow}
\usepackage{adjustbox}
\usepackage{booktabs}
\usepackage{color, colortbl}
\usepackage{bbding}
\usepackage{subcaption}
\usepackage{makecell}
\usepackage{algpseudocode}

\newcommand{\gc}{\cellcolor[gray]{0.915}}

\begin{document}

\title{GI-NAS: Boosting Gradient Inversion Attacks Through Adaptive Neural Architecture Search}

\author{
Wenbo Yu\orcidlink{0009-0004-8077-9487},
Hao Fang\orcidlink{0009-0004-0271-6579},
Bin Chen\orcidlink{0000-0002-4798-230X}, ~\IEEEmembership{Member, IEEE},
Xiaohang Sui\orcidlink{0009-0002-3653-5894},
Chuan Chen\orcidlink{0000-0002-7048-3445}, ~\IEEEmembership{Member, IEEE}, \\
Hao Wu\orcidlink{0009-0000-9034-4330},
Shu-Tao Xia\orcidlink{0000-0002-8639-982X}, ~\IEEEmembership{Member, IEEE},
Ke Xu\orcidlink{0000-0003-2587-8517}, ~\IEEEmembership{Fellow, IEEE}

\thanks{This work is supported in part by the National Science Foundation for Distinguished Young Scholars of China under No. 62425201, the National Natural Science Foundation of China under grant 62171248, 62301189, 62176269, National Key R\&D Program of China (2023YFB2703700), and Shenzhen Science and Technology Program under Grant KJZD20240903103702004, JCYJ20220818101012025, GXWD20220811172936001. \textit{(Wenbo Yu and Hao Fang contributed equally to this paper.)} \textit{(Corresponding author: Bin Chen.)}}
\thanks{Wenbo Yu, Hao Fang, and Shu-Tao Xia are with the Tsinghua Shenzhen International Graduate School, Tsinghua University, Shenzhen, Guangdong 518055, China (e-mail: \{wenbo.research, ffhibnese\}@gmail.com; xiast@sz.tsinghua.edu.cn).}
\thanks{Bin Chen and Xiaohang Sui are with the School of Computer Science and Technology, Harbin Institute of Technology, Shenzhen, Guangdong 518055, China (e-mail: chenbin2021@hit.edu.cn; suixiaohang@stu.hit.edu.cn).}
\thanks{Chuan Chen is with the School of Computer Science and Engineering, Sun Yat-sen University, Guangzhou, Guangdong 510006, China (e-mail: chenchuan@mail.sysu.edu.cn).}
\thanks{Hao Wu is with the Shenzhen ShenNong Information Technology Co., Ltd., Shenzhen, Guangdong 518000, China, and also with the Tsinghua Shenzhen International Graduate School, Tsinghua University, Shenzhen, Guangdong 518055, China (e-mail: wu-h22@mails.tsinghua.edu.cn).}
\thanks{Ke Xu is with the Department of Computer Science and Technology, Tsinghua University, Beijing 100084, China (e-mail: xuke@tsinghua.edu.cn).}
\thanks{The source code is available at \textcolor{magenta}{\url{https://github.com/cswbyu/GI-NAS}}.}
}

\markboth{IEEE Transactions on Information Forensics and Security}{IEEE Transactions on Information Forensics and Security}


\maketitle

\begin{abstract}
Gradient Inversion Attacks invert the transmitted gradients in Federated Learning (FL) systems to reconstruct the sensitive data of local clients and have raised considerable privacy concerns. A majority of gradient inversion methods rely heavily on explicit prior knowledge (e.g., a well pre-trained generative model), which is often unavailable in realistic scenarios. This is because real-world client data distributions are often highly heterogeneous, domain-specific, and unavailable to attackers, making it impractical for attackers to obtain perfectly matched pre-trained models, which inevitably suffer from fundamental distribution shifts relative to target private data. To alleviate this issue, researchers have proposed to leverage the implicit prior knowledge of an over-parameterized network. However, they only utilize a fixed neural architecture for all the attack settings. This would hinder the adaptive use of implicit architectural priors and consequently limit the generalizability. In this paper, we further exploit such implicit prior knowledge by proposing Gradient Inversion via Neural Architecture Search (GI-NAS), which adaptively searches the network and captures the implicit priors behind neural architectures. Extensive experiments verify that our proposed GI-NAS can achieve superior attack performance compared to state-of-the-art gradient inversion methods, even under more practical settings with high-resolution images, large-sized batches, and advanced defense strategies. To the best of our knowledge, we are the first to successfully introduce NAS to the gradient inversion community. We believe that this work exposes critical vulnerabilities in real-world federated learning by demonstrating high-fidelity reconstruction of sensitive data without requiring domain-specific priors, forcing urgent reassessment of FL privacy safeguards.
\end{abstract}

\section{Introduction}

\IEEEPARstart{F}{ederated} Learning (FL) \cite{mcmahan2017communication,wei2020federated,zhang2021survey} serves as an efficient collaborative learning framework where multiple participants cooperatively train a global model and only the computed gradients are exchanged. By adopting this distributed paradigm, FL systems fully leverage the huge amounts of data partitioned across various clients for enhanced model efficacy and tackle the separateness of data silos \cite{li2022federated}. Moreover, since merely the gradients instead of the private data are uploaded to the server, the user privacy seems to be safely guaranteed as the private data is only available at the client side.

However, FL systems are actually not so secure as what people have expected \cite{melis2019exploiting,nasr2019comprehensive,lyu2020threats,luo2021feature,zhu2019deep,geiping2020inverting,li2022auditing,fang2023gifd,zhang2023generative}. Extensive studies have discovered that even the transmitted gradients can disclose the sensitive information of users. Early works \cite{melis2019exploiting,nasr2019comprehensive,lyu2020threats,luo2021feature} involve inferring the existence of certain samples in the dataset (i.e., Membership Inference Attacks \cite{saeidian2021quantifying}) or further revealing some properties of the private training set (i.e., Property Inference Attacks \cite{ganju2018property}) from the uploaded gradients. But unlike the above inference attacks that only partially reveal limited information of the private data, Gradient Inversion Attacks \cite{zhu2019deep,geiping2020inverting,li2022auditing,fang2023gifd,zhang2023generative} stand out as a more threatening privacy risk as they can completely reconstruct the sensitive data by inverting the gradients.

Zhu \textit{et al.} \cite{zhu2019deep} first formulate gradient inversion as an optimization problem and recover the private training images by minimizing the gradient matching loss (i.e., the distance between the dummy gradients and the real gradients) with image pixels regarded as trainable parameters. Ensuing works improve the attack performance on the basis of Zhu \textit{et al.} \cite{zhu2019deep} by designing label extraction techniques \cite{zhao2020idlg}, switching the distance metric and introducing regularization \cite{geiping2020inverting}, or considering settings with larger batch sizes \cite{yin2021see}, but still restrict their optimizations in the pixel space. To fill this gap, recent studies \cite{li2022auditing, jeon2021gradient, fang2023gifd} propose to explore various search algorithms within the Generative Adversarial Networks (GAN) \cite{goodfellow2020generative, creswell2018generative} to leverage the rich prior knowledge encoded in the pre-trained generative models. While incorporating these explicit priors indeed improves the attack performance, it is usually tough to pre-prepare such prerequisites in realistic FL scenarios where the data distribution at the client side is likely to be complex or unknown to the attackers. Therefore, Zhang \textit{et al.} \cite{zhang2023generative} propose to employ an over-parameterized convolutional network for gradient inversion and directly optimize the excessive network parameters on a fixed input, which does not require any explicit prior information but still outperforms GAN-based attacks. The reason behind this is that the structure of a convolutional network naturally captures image statistics prior to any learning \cite{ulyanov2018deep} and Zhang \textit{et al.} \cite{zhang2023generative} leverage this characteristic as implicit prior knowledge. However, Zhang \textit{et al.} \cite{zhang2023generative} only utilize a fixed network for all the attack settings, regardless of the specific batch to recover. An intuitive question naturally arises: \textit{Can we adaptively search the architecture of the over-parameterized network to better capture the implicit prior knowledge for each reconstructed batch?}

\begin{figure*}[tbp]
\centering
\begin{subfigure}{0.3\textwidth}
    \centering
    \includegraphics[width=\linewidth]{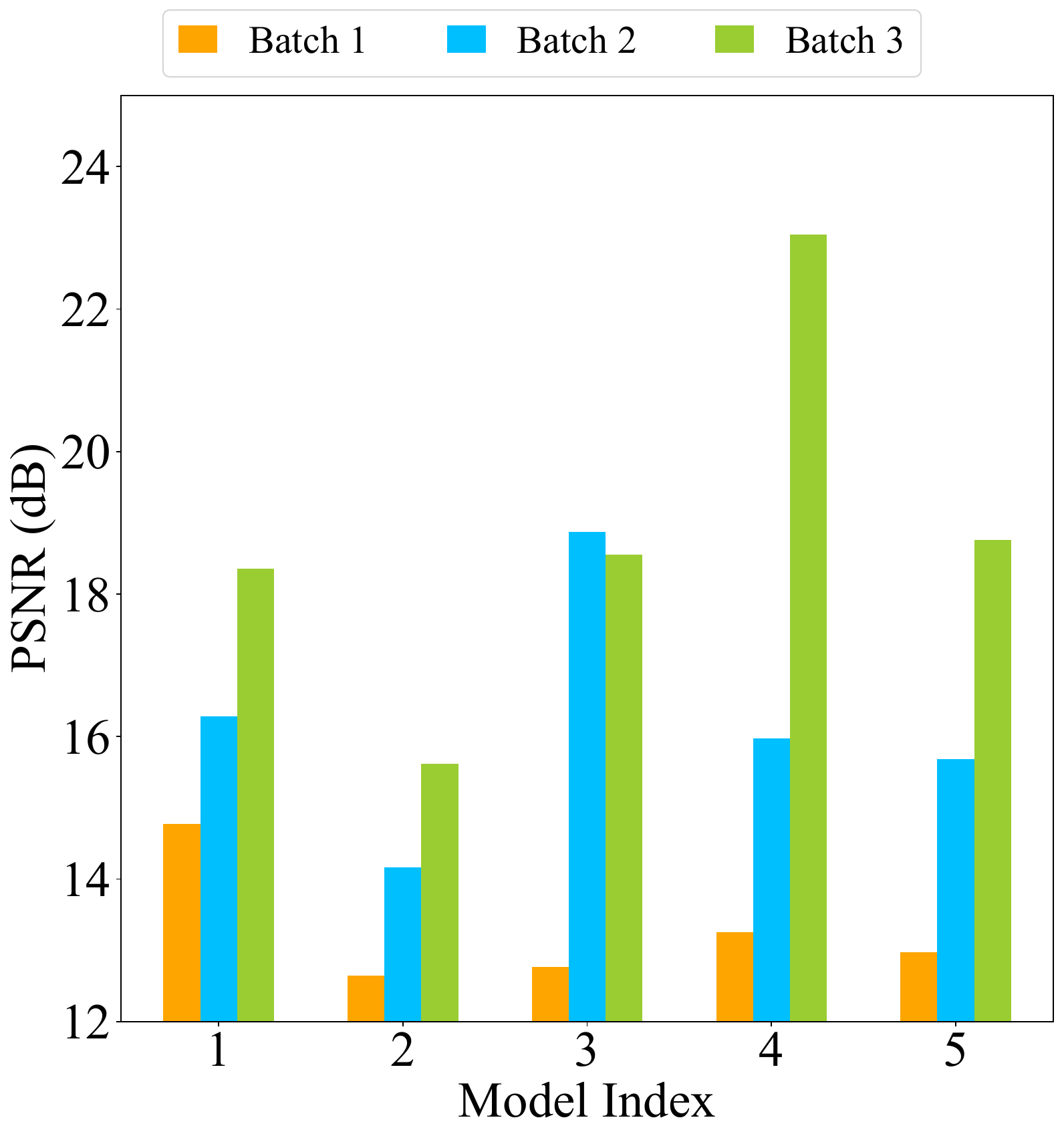}

    \caption{Quantitative Results}

\end{subfigure}
\centering
\begin{subfigure}{0.65\linewidth}
    \centering

    \begin{minipage}[t]{0.1575\linewidth}
    \centering
    \includegraphics[width=1.95cm]{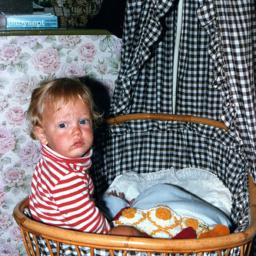}
    \centering
    \end{minipage}
    \begin{minipage}[t]{0.1575\linewidth}
    \centering
    \includegraphics[width=1.95cm]{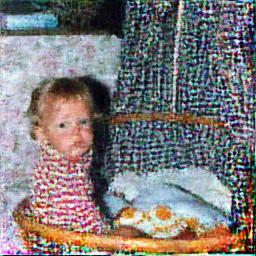}
    \centering
    \end{minipage}
    \begin{minipage}[t]{0.1575\linewidth}
    \centering
    \includegraphics[width=1.95cm]{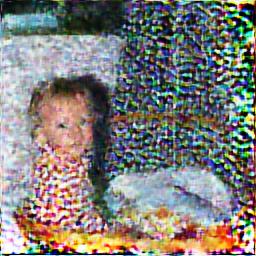}
    \centering
    \end{minipage}
    \begin{minipage}[t]{0.1575\linewidth}
    \centering
    \includegraphics[width=1.95cm]{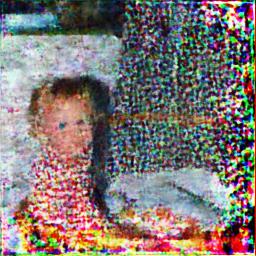}
    \centering
    \end{minipage}
    \begin{minipage}[t]{0.1575\linewidth}
    \centering
    \includegraphics[width=1.95cm]{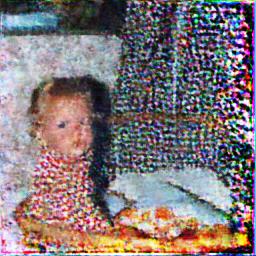}
    \centering
    \end{minipage}
    \begin{minipage}[t]{0.1575\linewidth}
    \centering
    \includegraphics[width=1.95cm]{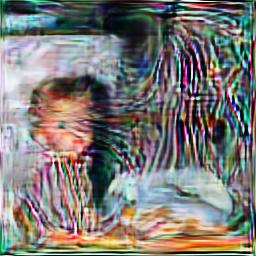}
    \centering
    \end{minipage}

    \vspace{0.9pt}

    \begin{minipage}[t]{0.1575\linewidth}
    \centering
    \includegraphics[width=1.95cm]{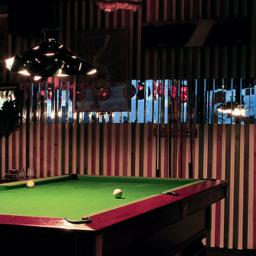}
    \centering
    \end{minipage}
    \begin{minipage}[t]{0.1575\linewidth}
    \centering
    \includegraphics[width=1.95cm]{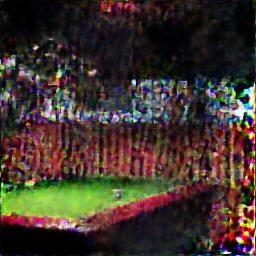}
    \centering
    \end{minipage}
    \begin{minipage}[t]{0.1575\linewidth}
    \centering
    \includegraphics[width=1.95cm]{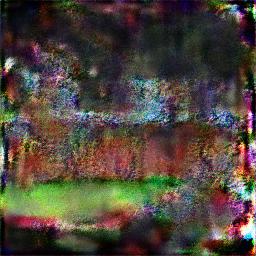}
    \centering
    \end{minipage}
    \begin{minipage}[t]{0.1575\linewidth}
    \centering
    \includegraphics[width=1.95cm]{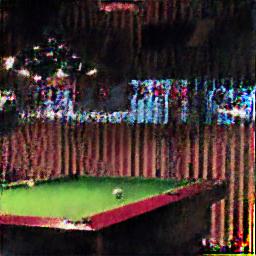}
    \centering
    \end{minipage}
    \begin{minipage}[t]{0.1575\linewidth}
    \centering
    \includegraphics[width=1.95cm]{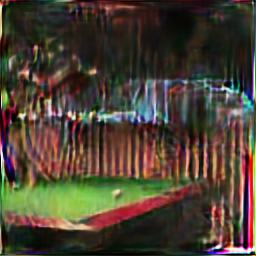}
    \centering
    \end{minipage}
    \begin{minipage}[t]{0.1575\linewidth}
    \centering
    \includegraphics[width=1.95cm]{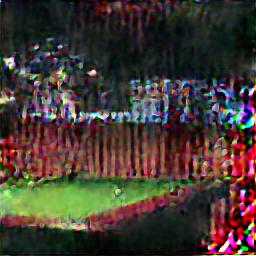}
    \centering
    \end{minipage}

    \vspace{0.9pt}

    \begin{minipage}[t]{0.1575\linewidth}
    \centering
    \includegraphics[width=1.95cm]{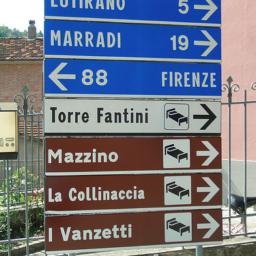}
    \caption*{\textbf{Original}}
    \centering
    \end{minipage}
    \begin{minipage}[t]{0.1575\linewidth}
    \centering
    \includegraphics[width=1.95cm]{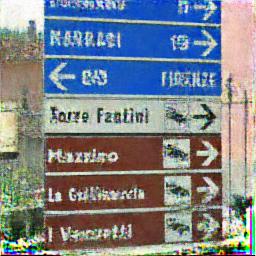}
    \caption*{\textbf{Model $1$}}
    \centering
    \end{minipage}
    \begin{minipage}[t]{0.1575\linewidth}
    \centering
    \includegraphics[width=1.95cm]{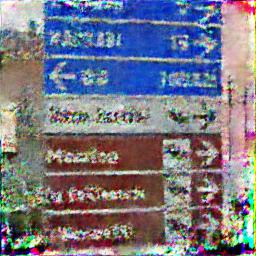}
    \caption*{\textbf{Model $2$}}
    \centering
    \end{minipage}
    \begin{minipage}[t]{0.1575\linewidth}
    \centering
    \includegraphics[width=1.95cm]{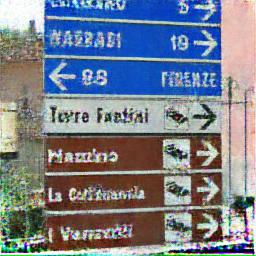}
    \caption*{\textbf{Model $3$}}
    \centering
    \end{minipage}
    \begin{minipage}[t]{0.1575\linewidth}
    \centering
    \includegraphics[width=1.95cm]{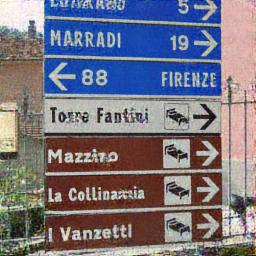}
    \caption*{\textbf{Model $4$}}
    \centering
    \end{minipage}
    \begin{minipage}[t]{0.1575\linewidth}
    \centering
    \includegraphics[width=1.95cm]{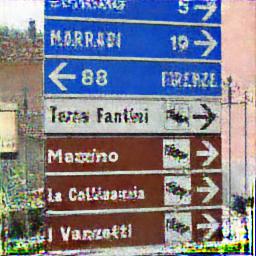}
    \caption*{\textbf{Model $5$}}
    \centering
    \end{minipage}

    \caption{Qualitative Results}
    \label{fig:introduction_quantitative}

\end{subfigure}
\centering
\caption{Quantitative and qualitative results when randomly attacking $3$ different batches by $5$ different models on ImageNet. In Figure \ref{fig:introduction_quantitative}, the first, second, and third rows are respectively from Batch $1$, Batch $2$, and Batch $3$.}
\label{fig:introduction}
\end{figure*}

In Figure \ref{fig:introduction}, we randomly select $5$ over-parameterized networks with different architectures to attack $3$ different batches. All the networks are optimized for the same number of iterations on ImageNet \cite{deng2009imagenet}. The results indicate that the Peak Signal-to-Noise Ratio (PSNR) performance varies significantly when changing the architectures. For the same batch, various architectures can hold remarkably different implicit priors on it. Besides, since the optimal models for Batch $1$, Batch $2$, and Batch $3$ are respectively Model $1$, Model $3$, and Model $4$, there exists no network that can consistently perform the best on all the given batches. Thus, the rigid use of a single architecture by Zhang \textit{et al.} \cite{zhang2023generative} lacks optimality under dynamic FL scenarios where different batches require different implicit architectural priors, and it is of great significance to adaptively select the most suitable architecture for each batch.

Inspired by the above phenomena, we propose a novel gradient inversion method, named \textbf{G}radient \textbf{I}nversion via \textbf{N}eural \textbf{A}rchitecture \textbf{S}earch (GI-NAS), to better match each batch with the optimal model architecture. Specifically, we first enlarge the potential search space for the over-parameterized network by designing different upsampling modules and skip connection patterns. To reduce the computational overhead, we utilize a training-free search strategy that compares the initial gradient matching loss for a given batch over all the candidates and selects the best of them for the final optimization. We further provide substantial experimental evidence that such a metric highly correlates with the real attack performance. We also consider more rigorous and realistic scenarios where the victims may hold high-resolution images and large-sized batches for training, and evaluate advanced defense strategies. Extensive experiments validate that GI-NAS can achieve state-of-the-art performance compared to existing gradient inversion methods. To the best of our knowledge, we are the first to introduce Neural Architecture Search (NAS) to Gradient Inversion Attacks. Our main contributions are as follows:

\begin{itemize}
  \item We systematically analyze existing methods, emphasize the necessity of equipping image batch recovery with the optimal model structure, and propose GI-NAS to boost gradient inversion through neural architectural search.
  \item We expand the model search space by considering different upsampling units and skip connection modes, and utilize a training-free search that regards the initial gradient matching loss as the search metric. We also provide extensive experimental evidence that such a metric highly correlates with the real attack performance.
  \item Numerous experimental results demonstrate that GI-NAS outperforms state-of-the-art gradient inversion methods, even under extreme settings with high-resolution images, large-sized batches, and advanced defense strategies.
  \item We provide deeper analysis on various aspects, such as computational efficiency, ablation studies, generalizability to more FL global models, robustness to network parameters and latent codes initialization, and implications behind the NAS searched results.
\end{itemize}

\textbf{Research Purpose and Real-World Significance.} This research reveals a critical security vulnerability in federated learning by developing GI-NAS, an adaptive gradient inversion attack that dynamically adjusts to diverse real-world conditions. Unlike existing methods constrained by fixed architectures or unrealistic data assumptions, our approach automatically identifies optimal network configurations through neural architecture search, enabling more effective privacy attacks across varying batch characteristics. The practical significance lies in exposing security risks in actual FL deployments where data resolutions, batch sizes, and defense mechanisms constantly vary. By demonstrating how the architectural adaptation enhances the attack effectiveness, we believe that this work will not only advance attack methodologies but also provide crucial insights for developing stronger defense systems in real-world distributed learning environments where sensitive data protection remains paramount.
\section{Related Work}

\subsection{Gradient Inversion Attacks and Defenses}

Zhu \textit{et al.} \cite{zhu2019deep} first propose to restore the private samples via iterative optimization for pixel-level reconstruction, yet limited to low-resolution and single images. Geiping \textit{et al.} \cite{geiping2020inverting} empirically decompose the gradient vector by its norm magnitude and updating direction, and succeed on high-resolution ImageNet \cite{deng2009imagenet} through the angle-based loss design. Furthermore, Yin \textit{et al.} \cite{yin2021see} extend the attacks to the batch-level inversion through the group consistency regularization and the improved batch label inference algorithm \cite{zhao2020idlg}. With strong handcrafted explicit priors (e.g., fidelity regularization, BN statistics), they accurately realize batch-level reconstruction with detailed semantic features allocated to all the individual images in a batch. Subsequent studies \cite{li2022auditing, jeon2021gradient, fang2023gifd} leverage pre-trained GAN models as generative priors \cite{li2025rethinking, yu2024editable, chen2024editable, fang2024privacy, fang2024one, qiu2024closer, qiu2024mibench} to enhance the attacks. Current defenses focus on gradient perturbation to alleviate the impact of gradient inversion with degraded gradients \cite{huang2021evaluating, zhang2022survey}. Gaussian Noise \cite{geyer2017differentially} and Gradient Clipping \cite{wei2021gradient} are common techniques in Differential Privacy (DP) \cite{zhu2014correlated, wei2023personalized} that effectively constrain the attackers from learning through the released gradients. Gradient Sparsification \cite{aji2017sparse, lin2018deep} prunes the gradients through a given threshold, and Soteria \cite{sun2021soteria} edits the gradients from the aspect of learned representations.

\subsection{Neural Architecture Search (NAS)}

By automatically searching the optimal model architecture, NAS algorithms have shown significant effectiveness in multiple visual tasks such as image restoration \cite{suganuma2018exploiting, chu2021fast}, semantic segmentation \cite{nekrasov2019fast, liu2019auto}, and image classification \cite{zoph2017neural, qin2023ag}. In image classification tasks, Zoph \textit{et al.} \cite{zoph2017neural} regularize the search space to a convolutional cell and construct a better architecture stacked by these cells using an RNN \cite{wang2016survey} controller. For semantic segmentation tasks, Liu \textit{et al.} \cite{liu2019auto} search for the optimal network on the level of hierarchical network architecture and extend NAS to dense image prediction. In terms of image restoration tasks, Suganuma \textit{et al.} \cite{suganuma2018exploiting} exploit a better Convolutional Autoencoders (CAE) with standard network components through the evolutionary search, while Chu \textit{et al.} \cite{chu2021fast} discover a competitive lightweight model via both micro and macro architecture search.

Existing NAS methods adopt different strategies to exploit the search space, including evolutionary methods \cite{liu2017hierarchical, miikkulainen2024evolving, jozefowicz2015empirical}, Bayesian optimization \cite{bergstra2013making, mendoza2016towards, domhan2015speeding}, Reinforcement Learning (RL) \cite{baker2016designing, cai2018efficient, zhong2018practical, zoph2017neural}, and gradient-based search \cite{xie2018snas, cai2018proxylessnas, ahmed2018maskconnect}. Different RL-based methods vary in the way to represent the agent's policy and the optimization process. Zoph \textit{et al.} \cite{zoph2017neural} utilize RNN networks to sequentially encode the neural architecture and train them with the REINFORCE policy gradient algorithm \cite{williams1992simple}. Baker \textit{et al.} \cite{baker2016designing} adopt Q-learning \cite{watkins1992q, clifton2020q} to train the policy network and realize competitive model design. Notably, recent works \cite{zhou2022training, zhou2024training, serianni2023training} propose training-free NAS to mitigate the issue of huge computational expenses. Instead of training from scratch, they evaluate the searched networks by some empirically designed metrics that reflect model effectiveness. In this paper, we adopt the initial gradient matching loss as the training-free search metric and provide substantial empirical evidence that such a search metric highly correlates with the real attack performance.
\section{Problem Formulation}

\textbf{Basics of Gradient Inversion.} We consider the training process of a classification model $f_\theta$ parameterized by $\theta$ in FL scenarios. The real gradients $\mathbf{g}$ are calculated from a private batch (with real images $\mathbf{x}$ and real labels $\mathbf{y}$) at the client side. The universal goal of Gradient Inversion Attacks is to search for some fake images $\mathbf{\hat{x}} \in \mathbb{R}^{B \times H \times W \times C}$ with labels $\mathbf{\hat{y}} \in \{0, 1\}^{B \times L}$ so that ($\mathbf{\hat{x}}$, $\mathbf{\hat{y}}$) can be close to ($\mathbf{x}$, $\mathbf{y}$) as much as possible, where $B$, $H$, $W$, $C$, and $L$ are respectively batch size, image height, image width, number of channels, and number of classes. This can be realized by minimizing the gradient matching loss \cite{zhu2019deep}:

\begin{equation}
\label{eq:original_formulation}
\begin{aligned}
    \mathbf{\hat{x}^{*}}, \mathbf{\hat{y}^{*}} = \mathop{\arg\min}_{\mathbf{\hat{x}}, \mathbf{\hat{y}}} \mathcal{D}(\nabla_{\theta}\mathcal{L}(f_\theta(\mathbf{\hat{x}}), \mathbf{\hat{y}}), \mathbf{g}),
\end{aligned}
\end{equation}
\noindent where $\mathcal{D(\cdot, \cdot)}$ is the distance metric (e.g., $l_2$-norm loss, cosine-similarity loss) for the gradient matching loss and $\mathcal{L}(\cdot, \cdot)$ is the loss function of the global model $f_\theta$.

Previous works \cite{zhao2020idlg, yin2021see} in this field have revealed that the ground truth labels $\mathbf{y}$ can be directly inferred from the uploaded gradients $\mathbf{g}$. Therefore, the formulation in (\ref{eq:original_formulation}) can be simplified as:

\begin{equation}
\label{eq:simplified_formulation}
\begin{aligned}
    \mathbf{\hat{x}^{*}} = \mathop{\arg\min}_{\mathbf{\hat{x}}} \mathcal{D}(F(\mathbf{\hat{x}}), \mathbf{g}),
\end{aligned}
\end{equation}

\noindent where $F(\mathbf{\hat{x}})=\nabla_{\theta}\mathcal{L}(f_\theta(\mathbf{\hat{x}}), \mathbf{\hat{y}})$ calculates the gradients of $f_\theta$ provided with $\mathbf{\hat{x}}$.

The key challenge of (\ref{eq:simplified_formulation}) is that gradients only provide limited information of private data and there even exists a pair of different data having the same gradients \cite{zhu2020r}. To mitigate this issue, subsequent works incorporate various regularizations (e.g., total variation loss \cite{geiping2020inverting}, group consistency loss \cite{yin2021see}) as prior knowledge. Thus, the overall optimization becomes:

\begin{equation}
\label{eq:simplified_formulation_regularization}
\begin{aligned}
    \mathbf{\hat{x}^{*}} = \mathop{\arg\min}_{\mathbf{\hat{x}}} \mathcal{D}(F(\mathbf{\hat{x}}), \mathbf{g}) + \lambda\mathcal{R}_{prior}(\mathbf{\hat{x}}),
\end{aligned}
\end{equation}

\noindent where $\mathcal{R}_{prior}(\cdot)$ is the introduced regularization that can establish some priors for the attacks, and $\lambda$ is the weight factor.

\begin{figure*}[t]
\centerline{\includegraphics[width=\textwidth]{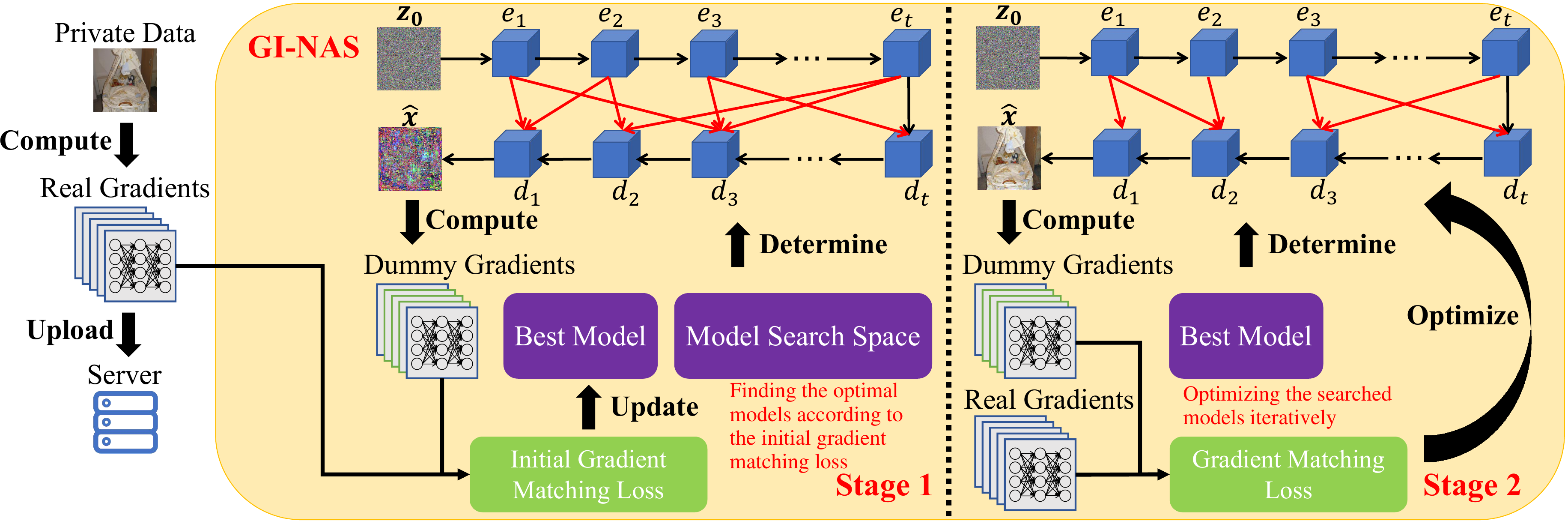}}
\caption{Overview of the proposed GI-NAS attack. We leverage a two-stage strategy for private batch recovery. In the first stage, we traverse the model search space and calculate the initial gradient matching loss (i.e., our training-free search metric) of each model based on the fixed input $\mathbf{z_0}$. We regard the model that achieves the minimal initial loss as our best model, for its performance at the start can stand out from numerous candidates. In the second stage, we adopt the architecture of the previously found best model and optimize its excessive parameters to reconstruct the private data.}
\label{fig:pipeline}
\end{figure*}

\textbf{GAN-based Gradient Inversion.} Nevertheless, the optimization of (\ref{eq:simplified_formulation_regularization}) is still limited in the pixel space. Given a well pre-trained GAN, an instinctive idea is to switch the optimization from the pixel space to the GAN latent space:

\begin{equation}
\label{eq:gan_formulation_regularization}
\begin{aligned}
    \mathbf{z^{*}} = \mathop{\arg\min}_{\mathbf{z}} \mathcal{D}(F(G_{\omega}(\mathbf{z})), \mathbf{g}) + \lambda\mathcal{R}_{prior}(G_{\omega}(\mathbf{z})),
\end{aligned}
\end{equation}

\noindent where $G_{\omega}$ and $\mathbf{z} \in \mathbb{R}^{B \times l}$ are respectively the generator and the latent vector of the pre-trained GAN. By reducing the optimization space from $\mathbb{R}^{B \times H \times W \times C}$ to $\mathbb{R}^{B \times l}$, (\ref{eq:gan_formulation_regularization}) overcomes the uncertainty of directly optimizing the extensive pixels and exploits the abundant prior knowledge encoded in the pre-trained GAN. Based on this, recently emerged GAN-based attacks \cite{li2022auditing, jeon2021gradient, fang2023gifd} explore various search strategies within the pre-trained GAN to utilize its expression ability.

\textbf{Gradient Inversion via Over-parameterized Networks.} But as previously mentioned, incorporating such GAN priors is often impractical in realistic scenarios where the distribution of $\mathbf{x}$ is likely to be mismatched with the training data of the pre-trained GAN. Furthermore, it has already been discovered in \cite{zhang2023generative} that explicitly introducing regularization in (\ref{eq:simplified_formulation_regularization}) or (\ref{eq:gan_formulation_regularization}) may not necessarily result in convergence towards $\mathbf{x}$, as even ground truth images cannot guarantee minimal loss when $\mathcal{R}_{prior}(\cdot)$ is added. To mitigate these issues, Zhang \textit{et al.} \cite{zhang2023generative} propose to leverage an over-parameterized network as implicit prior knowledge:

\begin{equation}
\label{eq:over_parameterized_formulation}
\begin{aligned}
    \mathbf{\phi^{*}} = \mathop{\arg\min}_{\mathbf{\phi}} \mathcal{D}(F(G_{over}(\mathbf{z_0};\mathbf{\phi})), \mathbf{g}),
\end{aligned}
\end{equation}

\noindent where $G_{over}$ is the over-parameterized convolutional network with excessive parameters $\mathbf{\phi}$, and $\mathbf{z_0}$ is the randomly generated but fixed latent code. Note that the regularization term is omitted in (\ref{eq:over_parameterized_formulation}). This is because the architecture of $G_{over}$ itself can serve as implicit regularization, for convolutional networks have been discovered to possess implicit priors that prioritize clean images rather than noise as shown in \cite{ulyanov2018deep}. Thus, the generated images that highly resemble the ground truth images can be obtained through $\mathbf{\hat{x}^{*}} = G_{over}(\mathbf{z_0};\mathbf{\phi^{*}})$. However, only a changeless over-parameterized network is employed for all the attack settings in \cite{zhang2023generative}. As previously shown in Figure \ref{fig:introduction}, although the network is over-parameterized, the attack performance exhibits significant differences when adopting different architectures. To fill this gap, we propose to further exploit such implicit architectural priors by searching the optimal over-parameterized network $G_{opt}$ for each batch. We will discuss how to realize this in Section \ref{sec:gi-nas}.
\section{Method}
\label{sec:gi-nas}

Our proposed GI-NAS attack is carried out in two stages. In the first stage, we conduct our architecture search to decide the optimal model $G_{opt}$. Given that the attackers may only hold limited resources, we utilize the initial gradient matching loss as the training-free search metric to reduce the computational overhead. In the second stage, we iteratively optimize the parameters of the selected model $G_{opt}$ to recover the sensitive data. Figure \ref{fig:pipeline} illustrates the overview of our method.

\subsection{Threat Model}

\textbf{Attacker's Aim.} The fundamental aim of an attacker in our method is to reconstruct the original private training data $\mathbf{x}$ from client devices with maximum fidelity by exploiting the exchanged gradients $\mathbf{g}$ in federated learning systems.

\textbf{Attacker's Knowledge.} The attacker possesses three key categories of knowledge: complete access to the gradient vectors $\mathbf{g}$ transmitted during federated updates, including their structural organization and numerical values; full architectural specifications of the global model $f_\theta$ being trained; and the standard assumptions about input data dimensions $\mathbf{x} \in \mathbb{R}^{B \times H \times W \times C}$ without requiring distributional priors that are needed in previous GAN-based attacks \cite{li2022auditing, fang2023gifd, jeon2021gradient}. Crucially, the knowledge requirement in our method excludes any direct access to the original training data samples or client-side information beyond the gradients, making our threat model both realistic and concerning for practical deployments.

\textbf{Attacker's Capabilities.} The attacker's capabilities encompass executing local computations to optimize reconstruction networks $G_{opt}$, deploying adaptive architectures through our NAS framework to handle varying batch characteristics, and maintaining attack efficacy even when gradients are protected by common defense mechanisms. The attacker is not allowed to modify the exchanged gradients or transmit malicious data that may compromise the training process, even if doing so can enhance the reconstruction results.

\subsection{Stage 1: Training-free Optimal Model Search}
\label{sec:stage_1}

Before we elaborate on the Stage 1 (i.e., Training-free Optimal Model Search), we outline its key components and objectives as follows:

\begin{itemize}
    \item Input for this stage: Real gradients $\mathbf{g}$, fixed latent code $\mathbf{z_0}$ sampled from $\mathcal{N}(0, 1)$, and the model search space $\mathcal{M} = \{G_{1}, G_{2}, ..., G_{n}\}$.
    \item Output for this stage: The searched optimal architecture $G_{opt}$ with minimal initial loss.
    \item Purpose for this stage: In this stage, we traverse the model search space and calculate the initial gradient matching loss (i.e., our training-free search metric) of each model based on the fixed input $\mathbf{z_0}$. We regard the model that achieves the minimal initial loss as our best model $G_{opt}$.
\end{itemize}

\begin{figure*}[t]
\centerline{\includegraphics[width=\textwidth]{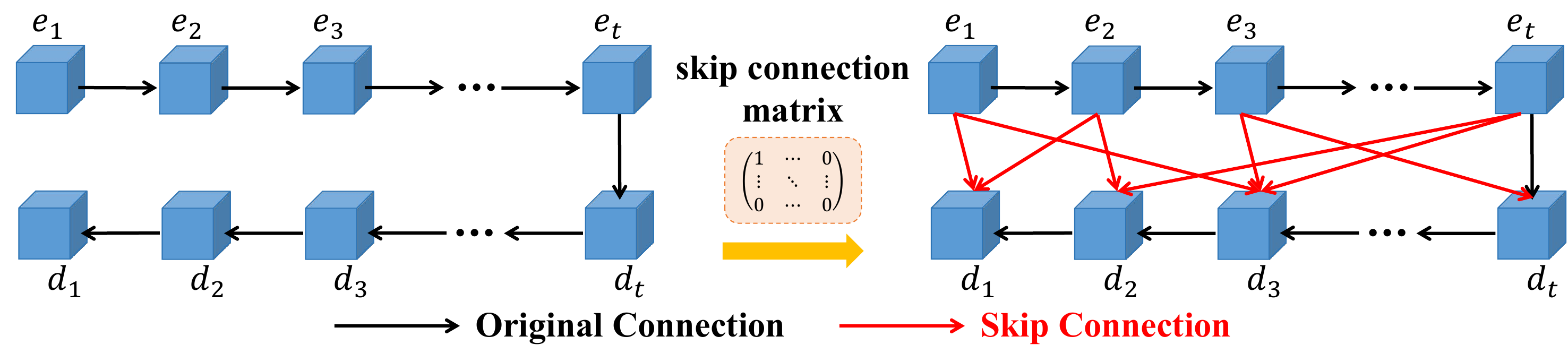}}
\caption{The design of search space for skip connection patterns. Different skip connection patterns are determined by the skip connection matrix $\mathbf{A} \in \{0, 1\}^{t \times t}$. $\mathbf{A}_{ij} = 1$ indicates that there exists a skip connection from $e_{i}$ to $d_{j}$ and $\mathbf{A}_{ij} = 0$ means that there is not such a skip connection.}
\label{fig:skip_connection_search_space}
\end{figure*}

\subsubsection{Model Search Space Design}

One crucial factor for NAS is that the potential search space is large and diverse enough to cover the optimal design. Therefore, we adopt U-Net \cite{ronneberger2015u}, a typical convolutional neural architecture as the fundamental of our model search, since the skip connection patterns between its encoders and decoders can provide adequate alternatives of model structure. Besides, the configurations of the upsampling modules (e.g., kernel size, activation function) can also enable numerous possibilities when arranged and combined. Similar to previous NAS methods \cite{chen2020dip, arican2022isnas}, we enlarge the search space of our model from two aspects, namely \textit{Upsampling Modules} and \textit{Skip Connection Patterns}.

\textbf{Search Space for Upsampling Modules.} We decompose the upsampling operations into five key components: feature upsampling, feature transformation, activation function, kernel size, and dilation rate. Then, we allocate a series of possible options to each of these components. When deciding on feature upsampling, we choose from commonly used interpolation techniques, such as bilinear interpolation, bicubic interpolation, and nearest-neighbour interpolation. As for feature transformation, we choose from classical convolution techniques, such as 2D convolution, separable convolution, and depth-wise convolution. As regards activation function, we select from ReLU, LeakyReLU, PReLU, etc. Furthermore, we supply kernel size and dilation rate with more choices, such as $1 \times 1$, $3 \times 3$, or $5 \times 5$ for kernel size and $1$, $3$, or $5$ for dilation rate. The combination of these flexible components can contribute to the diversity of upsampling modules.

\textbf{Search Space for Skip Connection Patterns.} We assume that there are $t$ levels of encoders and decoders in total, and denote them as $e_1, e_2, ..., e_t$ and $d_1, d_2, ..., d_t$. As shown in Figure \ref{fig:skip_connection_search_space}, We consider different skip connection patterns between encoders and decoders. To represent each of these patterns, we define a skip connection matrix $\mathbf{A} \in \{0, 1\}^{t \times t}$ that serves as a mask to determine whether there will be new residual connections \cite{he2016deep} between pairs of encoders and decoders. More concretely, $\mathbf{A}_{ij} = 1$ indicates that there exists a skip connection from $e_{i}$ to $d_{j}$ and $\mathbf{A}_{ij} = 0$ means that there is not such a skip connection. As the shapes of feature maps across different network levels can vary significantly (e.g., $64 \times 64$ for the output of $e_i$ and $256 \times 256$ for the input of $d_j$), we introduce connection scale factors to tackle this inconsistency and decompose all the possible scale factors into a series of $2\times$ upsampling operations or downsampling operations with shared weights. By allocating $0$ or $1$ to each of the $t^2$ bits in $\textbf{A}$, we broaden the search space of skip connection patterns as there are $2^{t^2}$ possibilities altogether and we only need to sample a portion of them.

\subsubsection{Optimal Model Selection}

We build up our model search space $\mathcal{M}$ by combining the possibilities of the aforementioned upsampling modules and skip connection patterns. We assume that the size of our model search space is $n$ and the candidates inside it are denoted as $\mathcal{M} = \{G_{1}, G_{2}, ..., G_{n}\}$. We first sample the latent code $\mathbf{z_0}$ from the Gaussian distribution and freeze its values on all the models for fair comparison. We then traverse the model search space and calculate the initial gradient matching loss of each individual $G_{r}\,(1 \leq r \leq n)$:

\begin{equation}
\label{eq:our_method}
\begin{aligned}
    \mathcal{L}_{grad}(G_{r}) = \mathcal{D}(\mathcal{T}(F(G_{r}(\mathbf{z_0};\mathbf{\phi}_{r}))), \mathbf{g}),
\end{aligned}
\end{equation}

\noindent where $\mathcal{L}_{grad}(\cdot)$ is the gradient matching loss, $\mathcal{T}(\cdot)$ is the estimated gradient transformation \cite{li2022auditing}, and $\mathbf{\phi}_{r}$ are the parameters of $G_{r}$. Here we introduce $\mathcal{T}(\cdot)$ to estimate the gradient transformation following the previous defense auditing work \cite{li2022auditing}, since the victims may apply defense strategies to $\mathbf{g}$ (e.g., Gradient Clipping \cite{geyer2017differentially}) and only release disrupted forms of gradients. Empirically, the model that can perform the best at the start and stand out from numerous candidates is likely to have better implicit architectural priors with respect to the private batch. This is because the model would have a superior optimization starting point (i.e., a smaller initial loss) due to its inherent structural advantages. We further provide extensive experimental evidence that such a metric highly correlates with the real attack performance in Section \ref{sec:further_analysis}. Therefore, we regard the model that achieves the minimal initial loss as our best model $G_{opt}$ and update our selection during the traversal. Since only the initial loss is calculated and no back-propagation is involved, this search process is training-free and hence computationally efficient.

\subsection{Stage 2: Private Batch Recovery via the Optimal Model}

Similar to Section \ref{sec:stage_1}, we outline the key components and objectives of this stage as follows:

\begin{itemize}
    \item Input for this stage: The searched optimal architecture $G_{opt}$ from Stage 1, the randomly initialized network parameters $\gamma_{1}$, real gradients $\mathbf{g}$, and the fixed latent code $\mathbf{z_0}$ from Stage 1.
    \item Output for this stage: The final parameters $\mathbf{\gamma^{*}}$, and the reconstructed private data $\mathbf{\hat{x}^{*}} = G_{opt}(\mathbf{z_0};\mathbf{\gamma^{*}})$.
    \item Purpose for this stage: In this stage, we iteratively optimize the parameters of the selected model $G_{opt}$ to recover the sensitive data.
\end{itemize}

After deciding on the optimal model $G_{opt}$ with the parameters $\gamma_{1}$, we iteratively conduct gradient decent optimizations on $G_{opt}$ to minimize the gradient matching loss:

\begin{equation}
\label{eq:our_method_update}
\begin{aligned}
    \mathbf{\gamma}_{k+1} = \mathbf{\gamma}_{k} - \eta\nabla_{\mathbf{\gamma}_{k}}\mathcal{L}_{grad}(G_{opt}),
\end{aligned}
\end{equation}

\noindent where $k$ is the current number of iterations $(1 \leq k \leq m)$, $\gamma_{k+1}$ are the parameters of $G_{opt}$ after the $k$-th optimization, and $\eta$ is the learning rate. Once the above process converges and we obtain $\mathbf{\gamma^{*}} = \mathbf{\gamma}_{m + 1}$ that satisfy the minimum loss, the private batch can be reconstructed by $\mathbf{\hat{x}^{*}} = G_{opt}(\mathbf{z_0};\mathbf{\gamma^{*}})$. In summary, the pseudocode of GI-NAS is illustrated in Algorithm \ref{algorithm}.

\begin{algorithm}[t]
    \caption{Gradient Inversion via Neural Architecture Search (GI-NAS)}
    \label{algorithm}
    \begin{algorithmic}[1]
        \Require
            $n$: the size of the model search space; $\mathbf{g}$: the uploaded real gradients; $m$: the maximum iteration steps for the final optimization;
        \Ensure
            $\mathbf{\hat{x}^{*}}$: the generated images;
        \State Build up the model search space: $\mathcal{M} = \{G_{1}, G_{2}, ..., G_{n}\}$ with the parameters $\Phi = \{\phi_{1}, \phi_{2}, ..., \phi_{n}\}$
        \State $\mathbf{z_0} \leftarrow \mathcal{N}(0, 1)$, $loss_{min} \leftarrow +\infty$
        \For {$i \leftarrow 1$ to $n$}
            \State $loss_{i} \leftarrow \mathcal{D}(\mathcal{T}(F(G_{i}(\mathbf{z_0};\mathbf{\phi}_{i}))), \mathbf{g})$ // calculate the initial gradient matching loss
            \If {$loss_{i} < loss_{min}$}
                \State $loss_{min} \leftarrow loss_{i}$, $G_{opt} \leftarrow G_{i}$, $\gamma_1 \leftarrow \mathbf{\phi}_{i}$ // update the selection of $G_{opt}$
            \EndIf
        \EndFor
        \For {$k \leftarrow 1$ to $m$}
            \State $\mathbf{\gamma}_{k+1} \leftarrow \mathbf{\gamma}_{k} - \eta\nabla_{\mathbf{\gamma}_{k}}\mathcal{D}(\mathcal{T}(F(G_{opt}(\mathbf{z_0};\mathbf{\gamma}_{k}))), \mathbf{g})$
        \EndFor
        \State $\mathbf{\gamma^{*}} \leftarrow \mathbf{\gamma}_{m + 1}$ \\
        \Return $\mathbf{\hat{x}^{*}} = G_{opt}(\mathbf{z_0};\mathbf{\gamma^{*}})$
    \end{algorithmic}
\end{algorithm}
\section{Experiments}

\subsection{Experimental Setup}

\textbf{Evaluation Settings.} We test our method on CIFAR-10 \cite{krizhevsky2009learning} and ImageNet \cite{deng2009imagenet} with the resolutions of $32 \times 32$ and $256 \times 256$. Unlike many previous methods \cite{jeon2021gradient, fang2023gifd} that scale down the ImageNet images to $64 \times 64$, here we emphasize that we adopt the high-resolution version of ImageNet. Thus, our settings are more rigorous and realistic. Following previous gradient inversion works \cite{zhu2019deep, zhang2023generative}, we adopt ResNet-$18$ \cite{he2016deep} as our global model and utilize the same preprocessing procedures. We build up the search space and randomly generate the alternative models by arbitrarily changing the options of upsampling modules and skip connection patterns.

\textbf{State-of-the-art Baselines for Comparison.} We implement the following gradient inversion baselines: (1) \textit{IG (Inverting Gradients)} \cite{geiping2020inverting}: pixel-level reconstruction with angle-based loss function; (2) \textit{GI (GradInversion)} \cite{yin2021see}: realizing batch-level restoration via multiple regularization priors; (3) \textit{GGL (Generative Gradient Leakage)} \cite{li2022auditing}: employing strong GAN priors to produce high-fidelity images under severe defense strategies; (4) \textit{GIAS (Gradient Inversion in Alternative Space)} \cite{jeon2021gradient}: searching the optimal latent code while optimizing in the generator parameter space; (5) \textit{GIFD (Gradient Inversion over Feature Domain)} \cite{fang2023gifd}: leveraging intermediate layer optimization for gradient inversion to further exploit pre-trained GAN priors; (6) \textit{GION (Gradient Inversion via Over-parameterized Networks)} \cite{zhang2023generative}: designing an over-parameterized convolutional network with excessive parameters and employing a fixed network architecture as implicit regularization. Note that when implementing GAN-based methods such as GGL and GIAS, we adopt BigGAN \cite{brock2018large} pre-trained on ImageNet, which may result in mismatched priors when the target data is from CIFAR-10. This is because there is an inherent distribution bias between the ImageNet and CIFAR-10 datasets. However, such mismatched priors can be very common under realistic FL scenarios as mentioned in previous works \cite{zhang2023generative, fang2023gifd}.

\textbf{Quantitative Metrics.} We utilize four metrics to measure the quality of reconstruction images: (1) \textit{Peak Signal-to-Noise Ratio (PSNR) $\uparrow$}; (2) \textit{Structural Similarity Index Measure (SSIM) $\uparrow$}; (3) \textit{Feature Similarity Index Measure (FSIM) $\uparrow$}; (4) \textit{Learned Perceptual Image Patch Similarity (LPIPS) $\downarrow$} \cite{zhang2018unreasonable}. Note that ``$\downarrow$" indicates that the lower the metric, the better the attack performance while ``$\uparrow$" indicates that the higher the metric, the better the attack performance.

\textbf{Implementation Details.} The learning rate $\eta$ in (\ref{eq:our_method_update}) is set as $1 \times 10^{-3}$. We utilize the signed gradient descent and adopt Adam optimizer \cite{kingma2014adam} when updating the parameters of $G_{opt}$. We choose the negative cosine similarity function as the distance metric $\mathcal{D(\cdot, \cdot)}$ when calculating the gradient matching loss $\mathcal{L}_{grad}(\cdot)$ in (\ref{eq:our_method}). We conduct all the experiments on NVIDIA GeForce RTX 3090 GPUs for smaller batch sizes and on A100 GPUs for larger batch sizes.

\subsection{Choices of search size $n$ and network depth $t$}

\begin{table*}[t]
    \caption{Attack results under different values of search size $n$. We randomly select images \textit{(strictly having no intersections with the images used in the remaining experiments of this paper)} from ImageNet with the default batch size $B = 4$ for testing.}
    \label{tab:choices_n}
    \centering
    \tabcolsep=2pt
    \adjustbox{width=\textwidth}{\begin{tabular}{ccccccccccccccc}
    \toprule
    Metric & 1 & 10 & 50 & 100 & 500 & 1000 & 2000 & 3000 & 4000 & \gc\textbf{5000} & 6000 & 7000 & 8000 & 9000 \\
    \midrule
    PSNR $\uparrow$ & 21.5034 & 21.6179 & 21.2776 & 22.0846 & 22.1451 & 22.5949 & 22.7380 & 23.3836 & 23.5626 & \gc\textbf{24.1603} & 23.1646 & 23.3276 & 23.2485 & 22.9561 \\
    SSIM $\uparrow$ & 0.6314 & 0.5960 & 0.6222 & 0.6069 & 0.6099 & 0.6335 & 0.5960 & 0.6314 & 0.6424 & \gc\textbf{0.6566} & 0.6518 & 0.6395 & 0.6324 & 0.6306 \\
    FSIM $\uparrow$ & 0.8281 & 0.7938 & 0.8019 & 0.8082 & 0.8125 & 0.8269 & 0.8028 & 0.8243 & 0.8337 & \gc\textbf{0.8439} & 0.8334 & 0.8347 & 0.8319 & 0.8228 \\
    LPIPS $\downarrow$ & 0.4391 & 0.5238 & 0.5212 & 0.4808 & 0.4839 & 0.4540 & 0.4499 & 0.4369 & 0.4317 & \gc\textbf{0.4103} & 0.4289 & 0.4281 & 0.4282 & 0.4913 \\
    \bottomrule
    \end{tabular}}
\end{table*}

\begin{table}[t]
    \caption{Attack results under different values of network depth $t$. We randomly select images \textit{(strictly having no intersections with the images used in the remaining experiments of this paper)} from ImageNet with the default batch size $B = 4$ for testing.}
    \label{tab:choices_t}
    \centering
    \tabcolsep=2pt
    \adjustbox{width=0.87\linewidth}{\begin{tabular}{cccccc}
    \toprule
    Metric & 3 & 4 & \gc\textbf{5} & 6 & 7 \\
    \midrule
    PSNR $\uparrow$ & 21.3108 & 22.4132 & \gc\textbf{24.1603} & 23.1865 & 22.9935 \\
    SSIM $\uparrow$ & 0.6114 & 0.6314 & \gc\textbf{0.6566} & 0.6441 & 0.6293 \\
    FSIM $\uparrow$ & 0.8249 & 0.8314 & \gc\textbf{0.8439} & 0.8317 & 0.8287 \\
    LPIPS $\downarrow$ & 0.4478 & 0.4263 & \gc\textbf{0.4103} & 0.4263 & 0.4301 \\
    \bottomrule
    \end{tabular}}
\end{table}

To obtain the best reconstruction performance, we need to carefully decide on the values of search size $n$ and network depth $t$. Note that both these two hyper-parameters reflect the degree of NAS, and there is a trade-off between under-fitting and over-fitting when choosing their values. Thus, we randomly select images \textit{(strictly having no intersections with the images used in the remaining experiments of this paper)} from ImageNet and test the attack results under different values of $n$ and $t$ with the default batch size $B = 4$. From Table \ref{tab:choices_n} and Table \ref{tab:choices_t}, we find that when $n$ and $t$ are small (e.g., $n = 100$ and $t = 3$), increasing any one of them indeed improves the attacks, as more optimal alternatives are provided or the models are more complicated to tackle the under-fitting. However, when $n$ and $t$ are large (e.g., $n = 7000$ and $t = 6$), further increasing any one of them cannot improve the attacks and will conversely lead to performance degradation due to the over-fitting. This is because candidates that focus too much on the minimal initial loss while neglecting the fundamental patterns of data recovery may be included in the search space and eventually selected. Thus, we adopt $n = 5000$ and $t = 5$ in the remaining experiments of this paper, as they strike a better balance and acquire the best results in Table \ref{tab:choices_n} and Table \ref{tab:choices_t}.

\subsection{Comparison with State-of-the-art Methods}
\label{sec:main_comparison}

\begin{table}[h]
    \caption{Quantitative comparison of GI-NAS to state-of-the-art gradient inversion methods on CIFAR-10 ($32 \times 32$) and ImageNet ($256 \times 256$) with the default batch size $B = 4$.}
    \label{tab:main_results}
    \centering
    \tabcolsep=2pt
    \adjustbox{width=\linewidth}{\begin{tabular}{ccccccccc}
    \toprule
    Dataset & Metric & IG & GI & GGL & GIAS & GIFD & GION & \gc\textbf{GI-NAS} \\
    \midrule
    \multirow{4}[0]{*}{CIFAR-10} & PSNR $\uparrow$ & 16.3188 & 15.4613 & 12.4938 & 17.3687 & 18.2325 & 30.8652 & \gc\textbf{35.9883} \\
    & SSIM $\uparrow$ & 0.5710 & 0.5127 & 0.3256 & 0.6239 & 0.6561 & 0.9918 & \gc\textbf{0.9983} \\
    & FSIM $\uparrow$ & 0.7564 & 0.7311 & 0.6029 & 0.7800 & 0.8025 & 0.9960 & \gc\textbf{0.9991} \\
    & LPIPS $\downarrow$ & 0.4410 & 0.4878 & 0.5992 & 0.4056 & 0.3606 & 0.0035 & \gc\textbf{0.0009} \\
    \midrule
    \multirow{4}[0]{*}{ImageNet} & PSNR $\uparrow$ & 7.9419 & 8.5070 & 11.6255 & 10.0602 & 9.8512 & 21.9942 & \gc\textbf{23.2578} \\
    & SSIM $\uparrow$ & 0.0815 & 0.1157 & 0.2586 & 0.2408 & 0.2475 & 0.6188 & \gc\textbf{0.6848} \\
    & FSIM $\uparrow$ & 0.5269 & 0.5299 & 0.5924 & 0.5719 & 0.5767 & 0.8198 & \gc\textbf{0.8513} \\
    & LPIPS $\downarrow$ & 0.7194 & 0.7168 & 0.6152 & 0.6563 & 0.6561 & 0.4605 & \gc\textbf{0.3952} \\
    \bottomrule
    \end{tabular}}
\end{table}

Firstly, we compare GI-NAS with state-of-the-art gradient inversion methods on CIFAR-10. From Table \ref{tab:main_results}, we conclude that GI-NAS achieves the best results with significant performance improvements. For instance, we realize a $5.12$ dB PSNR increase than GION, and our LPIPS value is $74.3\%$ smaller than that of GION. These validate that in contrast to GION that optimizes on a fixed network, our NAS strategy indeed comes into effect and better leverages the implicit architectural priors.

\begin{table*}[t]
    \caption{Quantitative comparison of GI-NAS to state-of-the-art gradient inversion methods on CIFAR-10 ($32 \times 32$) and ImageNet ($256 \times 256$) with larger batch sizes when $B > 4$.}
    \label{tab:larger_bs}
    \centering
    \tabcolsep=2pt
    \adjustbox{max width=\linewidth}{\begin{tabular}{cccccccccccccccccc}
    \toprule
    Dataset & Metric & Batch Size & IG & GI & GGL & GIAS & GIFD & GION & \gc\textbf{GI-NAS} & Batch Size & IG & GI & GGL & GIAS & GIFD & GION & \gc\textbf{GI-NAS} \\
    \midrule
    \multirow{8}[0]{*}{CIFAR-10} & PSNR $\uparrow$ & \multirow{4}[0]{*}{16} & 10.2647 & 11.2593 & 9.9964 & 11.1952 & 11.6853 & 24.8746 & \gc\textbf{30.5315} & \multirow{4}[0]{*}{32} & 9.5802 & 10.5863 & 9.4104 & 9.8669 & 11.0898 & 28.6832 & \gc\textbf{33.0586} \\
    & SSIM $\uparrow$ & & 0.2185 & 0.2431 & 0.1658 & 0.2638 & 0.3222 & 0.9816 & \gc\textbf{0.9948} & & 0.1595 & 0.2020 & 0.1481 & 0.1724 & 0.2824 & 0.9892 & \gc\textbf{0.9974} \\
    & FSIM $\uparrow$ & & 0.5254 & 0.5350 & 0.4988 & 0.5569 & 0.6393 & 0.9869 & \gc\textbf{0.9965} & & 0.5186 & 0.5325 & 0.4995 & 0.5180 & 0.6015 & 0.9882 & \gc\textbf{0.9984} \\
    & LPIPS $\downarrow$ & & 0.6254 & 0.6225 & 0.6655 & 0.5804 & 0.5925 & 0.0177 & \gc\textbf{0.0035} & & 0.6598 & 0.6626 & 0.6922 & 0.6362 & 0.6305 & 0.0218 & \gc\textbf{0.0017} \\
    \cmidrule(lr){2-18}
    & PSNR $\uparrow$ & \multirow{4}[0]{*}{48} & 9.5446 & 10.2000 & 9.1200 & 9.6376 & 10.3709 & 29.5312 & \gc\textbf{38.7054} & \multirow{4}[0]{*}{96} & 9.2363 & 9.9500 & 8.8575 & 9.2982 & 10.3227 & 28.9723 & \gc\textbf{31.8026} \\
    & SSIM $\uparrow$ & & 0.1574 & 0.1697 & 0.1552 & 0.1652 & 0.2460 & 0.9932 & \gc\textbf{0.9983} & & 0.1402 & 0.1643 & 0.1516 & 0.2180 & 0.2389 & 0.9878 & \gc\textbf{0.9941} \\
    & FSIM $\uparrow$ & & 0.5112 & 0.5167 & 0.4869 & 0.5126 & 0.5824 & 0.9949 & \gc\textbf{0.9987} & & 0.4961 & 0.5238 & 0.4801 & 0.5365 & 0.5775 & 0.9865 & \gc\textbf{0.9941} \\
    & LPIPS $\downarrow$ & & 0.6660 & 0.6571 & 0.7213 & 0.6554 & 0.6504 & 0.0072 & \gc\textbf{0.0020} & & 0.6865 & 0.6593 & 0.7248 & 0.7297 & 0.6476 & 0.0227 & \gc\textbf{0.0071} \\
    \midrule
    \multirow{8}[0]{*}{ImageNet} & PSNR $\uparrow$ & \multirow{4}[0]{*}{8} & 7.6011 & 8.3419 & 11.5454 & 9.6216 & 8.8377 & 20.4841 & \gc\textbf{20.8451} & \multirow{4}[0]{*}{16} & 7.7991 & 8.0233 & 11.4766 & 9.2563 & 8.8755 & 20.3078 & \gc\textbf{21.4267} \\
    & SSIM $\uparrow$ & & 0.0762 & 0.1113 & 0.2571 & 0.2218 & 0.2095 & 0.5681 & \gc\textbf{0.6076} & & 0.0704 & 0.0952 & 0.2561 & 0.2189 & 0.2193 & 0.5507 & \gc\textbf{0.6141} \\
    & FSIM $\uparrow$ & & 0.5103 & 0.5365 & 0.5942 & 0.5765 & 0.5726 & 0.7908 & \gc\textbf{0.8105} & & 0.5077 & 0.5165 & 0.5872 & 0.5832 & 0.5629 & 0.7840 & \gc\textbf{0.8157} \\
    & LPIPS $\downarrow$ & & 0.7374 & 0.7251 & 0.6152 & 0.6621 & 0.6701 & 0.5162 & \gc\textbf{0.4613} & & 0.7424 & 0.7362 & 0.6165 & 0.6656 & 0.6718 & 0.5313 & \gc\textbf{0.4613} \\
    \cmidrule(lr){2-18}
    & PSNR $\uparrow$ & \multirow{4}[0]{*}{24} & 7.1352 & 7.9079 & 10.9910 & 9.6601 & 9.2921 & 20.6845 & \gc\textbf{21.7933} & \multirow{4}[0]{*}{32} & 7.1085 & 7.8752 & 10.8987 & 10.0563 & 8.4252 & 16.1275 & \gc\textbf{20.0011} \\
    & SSIM $\uparrow$ & & 0.0549 & 0.0892 & 0.2583 & 0.2270 & 0.2421 & 0.5868 & \gc\textbf{0.6236} & & 0.0554 & 0.0763 & 0.2558 & 0.2361 & 0.2013 & 0.3415 & \gc\textbf{0.5485} \\
    & FSIM $\uparrow$ & & 0.4794 & 0.4943 & 0.5811 & 0.5618 & 0.5642 & 0.7978 & \gc\textbf{0.8163} & & 0.4940 & 0.4951 & 0.5889 & 0.5762 & 0.5545 & 0.6823 & \gc\textbf{0.7819} \\
    & LPIPS $\downarrow$ & & 0.7540 & 0.7475 & 0.6261 & 0.6667 & 0.6651 & 0.5219 & \gc\textbf{0.4651} & & 0.7493 & 0.7427 & 0.6296 & 0.6674 & 0.6853 & 0.6291 & \gc\textbf{0.5265} \\
    \bottomrule
    \end{tabular}}
\end{table*}

We also discover that the GAN-based method GGL underperforms the previous GAN-free methods (i.e., IG and GI). This is because GGL utilizes BigGAN \cite{brock2018large} pre-trained on ImageNet for generative priors in our settings, which has an inherent distribution bias with the target CIFAR-10 data domain. Besides, GGL only optimizes the latent vectors and cannot dynamically handle the mismatch between the training data of GAN and the target data. In contrast, GIAS optimizes both the latent vectors and the GAN parameters, which can reduce the distribution divergence and alleviate such mismatch to some extent. Therefore, although also GAN-based, GIAS exhibits much better performance than GGL on CIFAR-10.

\textbf{High-Resolution Images Recovery.} We then consider a more challenging situation and compare various methods on ImageNet with the resolution of $256\times 256$. In Table \ref{tab:main_results}, most methods encounter significant performance decline when attacking high-resolution images. The amplification of image pixels greatly increases the complexity of reconstruction tasks and thus obstructs the optimization process for optimal images. But GI-NAS still achieves the best attack results, with a PSNR increase of $1.26$ dB than GION.

\begin{figure*}[b]

        \begin{subfigure}{\linewidth}
        \centering
        \begin{minipage}[t]{0.1195\linewidth}
        \centering
        \includegraphics[width=2.245cm]{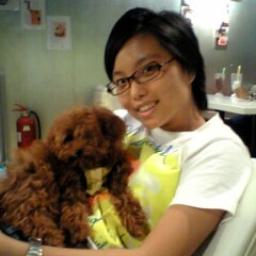}
        \centering
        \end{minipage}
        \begin{minipage}[t]{0.1195\linewidth}
        \centering
        \includegraphics[width=2.245cm]{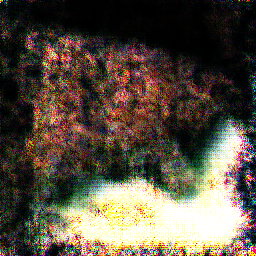}
        \centering
        \end{minipage}
        \begin{minipage}[t]{0.1195\linewidth}
        \centering
        \includegraphics[width=2.245cm]{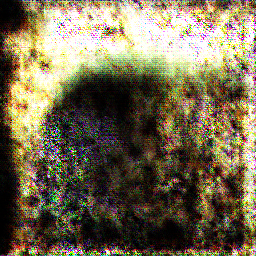}
        \centering
        \end{minipage}
        \begin{minipage}[t]{0.1195\linewidth}
        \centering
        \includegraphics[width=2.245cm]{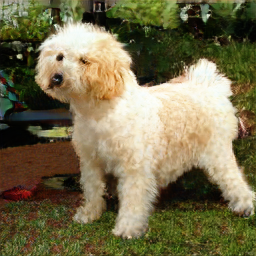}
        \centering
        \end{minipage}
        \begin{minipage}[t]{0.1195\linewidth}
        \centering
        \includegraphics[width=2.245cm]{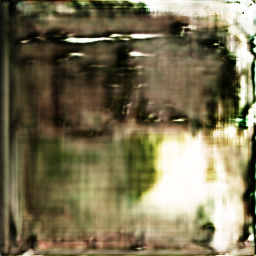}
        \centering
        \end{minipage}
        \begin{minipage}[t]{0.1195\linewidth}
        \centering
        \includegraphics[width=2.245cm]{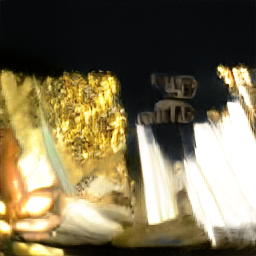}
        \centering
        \end{minipage}
        \begin{minipage}[t]{0.1195\linewidth}
        \centering
        \includegraphics[width=2.245cm]{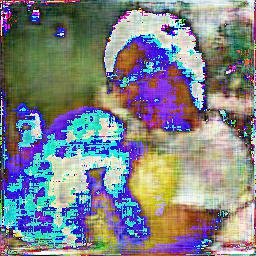}
        \centering
        \end{minipage}
        \begin{minipage}[t]{0.1195\linewidth}
        \centering
        \includegraphics[width=2.245cm]{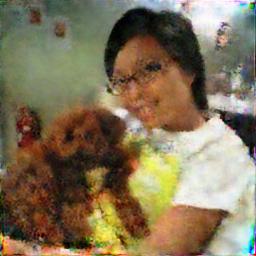}
        \centering
        \end{minipage}
        \end{subfigure}

        \vspace{0.9pt}

        \begin{subfigure}{\linewidth}
        \centering
        \begin{minipage}[t]{0.1195\linewidth}
        \centering
        \includegraphics[width=2.245cm]{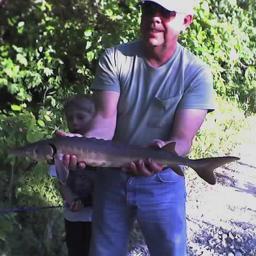}
        \centering
        \end{minipage}
        \begin{minipage}[t]{0.1195\linewidth}
        \centering
        \includegraphics[width=2.245cm]{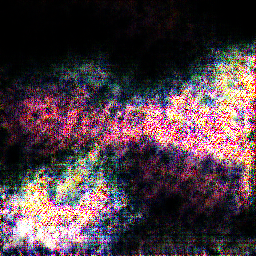}
        \centering
        \end{minipage}
        \begin{minipage}[t]{0.1195\linewidth}
        \centering
        \includegraphics[width=2.245cm]{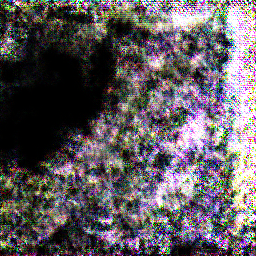}
        \centering
        \end{minipage}
        \begin{minipage}[t]{0.1195\linewidth}
        \centering
        \includegraphics[width=2.245cm]{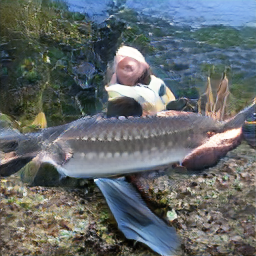}
        \centering
        \end{minipage}
        \begin{minipage}[t]{0.1195\linewidth}
        \centering
        \includegraphics[width=2.245cm]{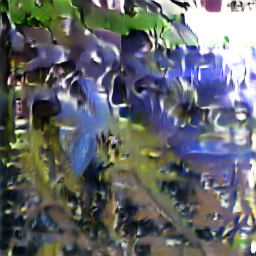}
        \centering
        \end{minipage}
        \begin{minipage}[t]{0.1195\linewidth}
        \centering
        \includegraphics[width=2.245cm]{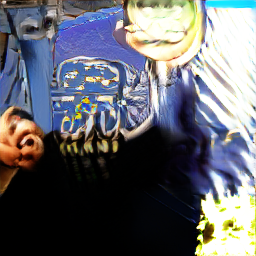}
        \centering
        \end{minipage}
        \begin{minipage}[t]{0.1195\linewidth}
        \centering
        \includegraphics[width=2.245cm]{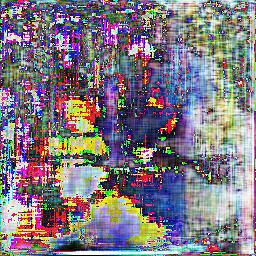}
        \centering
        \end{minipage}
        \begin{minipage}[t]{0.1195\linewidth}
        \centering
        \includegraphics[width=2.245cm]{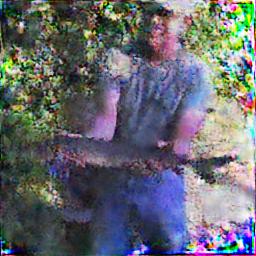}
        \centering
        \end{minipage}
        \end{subfigure}

        \vspace{0.9pt}

        \begin{subfigure}{\linewidth}
        \centering
        \begin{minipage}[t]{0.1195\linewidth}
        \centering
        \includegraphics[width=2.245cm]{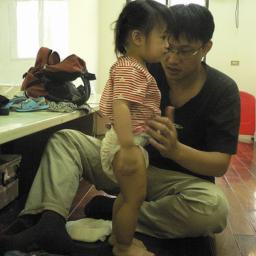}
        \caption*{\textbf{Original}}
        \centering
        \end{minipage}
        \begin{minipage}[t]{0.1195\linewidth}
        \centering
        \includegraphics[width=2.245cm]{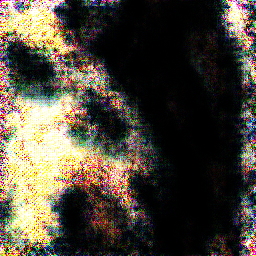}
        \caption*{\textbf{IG}}
        \centering
        \end{minipage}
        \begin{minipage}[t]{0.1195\linewidth}
        \centering
        \includegraphics[width=2.245cm]{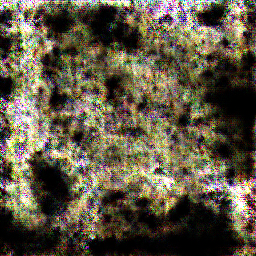}
        \caption*{\textbf{GI}}
        \centering
        \end{minipage}
        \begin{minipage}[t]{0.1195\linewidth}
        \centering
        \includegraphics[width=2.245cm]{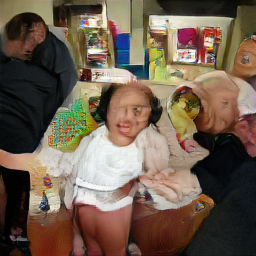}
        \caption*{\textbf{GGL}}
        \centering
        \end{minipage}
        \begin{minipage}[t]{0.1195\linewidth}
        \centering
        \includegraphics[width=2.245cm]{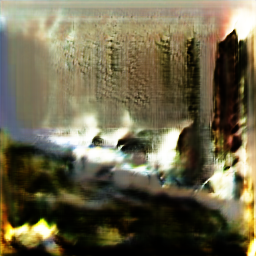}
        \caption*{\textbf{GIAS}}
        \centering
        \end{minipage}
        \begin{minipage}[t]{0.1195\linewidth}
        \centering
        \includegraphics[width=2.245cm]{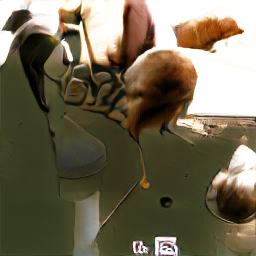}
        \caption*{\textbf{GIFD}}
        \centering
        \end{minipage}
        \begin{minipage}[t]{0.1195\linewidth}
        \centering
        \includegraphics[width=2.245cm]{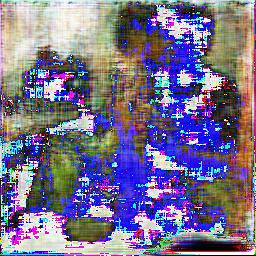}
        \caption*{\textbf{GION}}
        \centering
        \end{minipage}
        \begin{minipage}[t]{0.1195\linewidth}
        \centering
        \includegraphics[width=2.245cm]{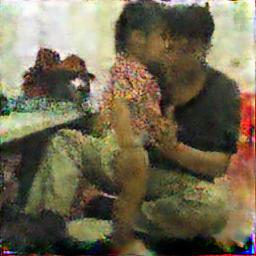}
        \caption*{\textbf{GI-NAS}}
        \centering
        \end{minipage}
        \end{subfigure}

    \caption{Qualitative comparison of GI-NAS to state-of-the-art gradient inversion methods on ImageNet ($256 \times 256$) with the larger batch size $B = 32$.}
    \label{visualization_imagenet}
\end{figure*}

\textbf{Extension to Larger Batch Sizes.} As shown in Table \ref{tab:larger_bs}, we extend GI-NAS to larger batch sizes both on CIFAR-10 and ImageNet, which is more in line with the actual training process of FL systems. We observe that the performance of most methods degrades as batch sizes increase because it would be harder for the attackers to distinguish each individuals in a batch, while GI-NAS is insusceptible to larger batch sizes and continuously generates high-quality images even at $B = 96$ on CIFAR-10 and $B = 32$ on ImageNet. Besides, GI-NAS is able to acquire consistent and significant performance gains on the basis of GION at all the given batch sizes. Figure \ref{visualization_imagenet} shows the qualitative comparison on ImageNet with the larger batch size $B = 32$. We notice that GI-NAS exactly realizes pixel-level recovery, while all the other compared methods struggle to perform the attacks and only obtain images with huge visual differences from the original ones. These results further provide evidence for the necessity and effectiveness of our batch-level optimal architecture search.

\textbf{Attacks under Defense Strategies.} Next, we evaluate how these attacks perform when defenses are applied both on CIFAR-10 and ImageNet. Following previous works \cite{li2022auditing, zhu2019deep}, we consider four strict defense strategies: (1) \textit{Gaussian Noise} \cite{geyer2017differentially} with a standard deviation of $0.1$; (2) \textit{Gradient Sparsification} \cite{aji2017sparse} with a pruning rate of $90\%$; (3) \textit{Gradient Clipping} \cite{geyer2017differentially} with a clipping bound of $4$; (4) \textit{Representative Perturbation (Soteria)} \cite{sun2021soteria} with a pruning rate of $80\%$. For fair comparison, we apply the estimated gradient transformation $\mathcal{T}(\cdot)$ described in (\ref{eq:our_method}) to all the attack methods. From Table \ref{tab:defense}, we discover that although the gradients have been disrupted by the imposed defense strategies, GI-NAS still realizes the best reconstruction effects in almost all the tested cases. The only exception is that GGL outperforms GI-NAS in terms of SSIM and LPIPS when the defense strategy is Gaussian Noise \cite{geyer2017differentially}. This is because the Gaussian noise with a standard deviation of $0.1$ can severely corrupt the gradients and the information carried inside the gradients is no longer enough for private batch recovery. However, GGL only optimizes the latent vectors and can still generate natural images that possess some semantic features by the pre-trained GAN. Thus, GGL can still obtain not bad performance even though the generated images are quite dissimilar to the original ones.

\begin{table*}[t]
    \caption{Quantitative comparison of GI-NAS to state-of-the-art gradient inversion methods on CIFAR-10 ($32 \times 32$) and ImageNet ($256 \times 256$) under various defense strategies with the default batch size $B = 4$.}
    \label{tab:defense}
    \centering
    \tabcolsep=2pt
    \adjustbox{max width=\linewidth}{\begin{tabular}{cccccccccccccccccc}
    \toprule
    Dataset & Metric & Defense & IG & GI & GGL & GIAS & GIFD & GION & \gc\textbf{GI-NAS} & Defense & IG & GI & GGL & GIAS & GIFD & GION & \gc\textbf{GI-NAS} \\
    \midrule
    \multirow{8}[0]{*}{CIFAR-10} & PSNR $\uparrow$ & \multirow{4}[0]{*}{\makecell{Gaussian \\ Noise}} & 8.8332 & 8.2797 & 10.3947 & 9.8146 & 9.8441 & 9.7658 & \gc\textbf{10.5192} & \multirow{4}[0]{*}{\makecell{Gradient \\ Sparsification}} & 9.6183 & 10.9486 & 10.4323 & 12.1328 & 14.0772 & 30.3518 & \gc\textbf{39.8739} \\
    & SSIM $\uparrow$ & & 0.1177 & 0.0976 & \textbf{0.1856} & 0.1301 & 0.1734 & 0.0280 & \gc0.0375 & & 0.1592 & 0.2168 & 0.1862 & 0.3227 & 0.4492 & 0.9920 & \gc\textbf{0.9987} \\
    & FSIM $\uparrow$ & & 0.4844 & 0.4806 & 0.5123 & 0.4843 & 0.5481 & 0.5945 & \gc\textbf{0.6300} & & 0.5151 & 0.5329 & 0.5223 & 0.6015 & 0.6984 & 0.9951 & \gc\textbf{0.9987} \\
    & LPIPS $\downarrow$ & & 0.7258 & 0.7077 & \textbf{0.6255} & 0.7129 & 0.7330 & 0.6752 & \gc0.6740 & & 0.6895 & 0.6855 & 0.6298 & 0.5582 & 0.5292 & 0.0047 & \gc\textbf{0.0014} \\
    \cmidrule(lr){2-18}
    & PSNR $\uparrow$ & \multirow{4}[0]{*}{\makecell{Gradient \\ Clipping}} & 16.6203 & 11.5888 & 10.1373 & 17.9859 & 19.0230 & 32.4744 & \gc\textbf{36.2490} & \multirow{4}[0]{*}{\makecell{Soteria}} & 9.4103 & 10.2939 & 10.4217 & 15.5825 & 17.6231 & 31.1003 & \gc\textbf{34.1804} \\
    & SSIM $\uparrow$ & & 0.5952 & 0.2851 & 0.1760 & 0.6703 & 0.7050 & 0.9933 & \gc\textbf{0.9985} & & 0.1404 & 0.1707 & 0.1889 & 0.5405 & 0.6319 & 0.9900 & \gc\textbf{0.9973} \\
    & FSIM $\uparrow$ & & 0.7464 & 0.4976 & 0.5070 & 0.7751 & 0.8237 & 0.9966 & \gc\textbf{0.9992} & & 0.5086 & 0.5086 & 0.5144 & 0.7133 & 0.7834 & 0.9952 & \gc\textbf{0.9986} \\
    & LPIPS $\downarrow$ & & 0.3811 & 0.6413 & 0.6205 & 0.3054 & 0.3313 & 0.0029 & \gc\textbf{0.0008} & & 0.7008 & 0.6940 & 0.6335 & 0.3748 & 0.3654 & 0.0038 & \gc\textbf{0.0014} \\
    \midrule
    \multirow{8}[0]{*}{ImageNet} & PSNR $\uparrow$ & \multirow{4}[0]{*}{\makecell{Gaussian \\ Noise}} & 7.7020 & 7.4553 & 9.7381 & 6.8520 & 8.3526 & 9.2449 & \gc\textbf{11.2567} & \multirow{4}[0]{*}{\makecell{Gradient \\ Sparsification}} & 8.6540 & 8.6519 & 9.8019 & 9.7894 & 9.8620 & 13.7358 & \gc\textbf{15.4331} \\
    & SSIM $\uparrow$ & & 0.0311 & 0.0197 & \textbf{0.2319} & 0.1045 & 0.1618 & 0.0277 & \gc0.0416 & & 0.0669 & 0.0627 & 0.2315 & 0.2515 & 0.2424 & 0.2387 & \gc\textbf{0.3154} \\
    & FSIM $\uparrow$ & & 0.4518 & 0.3804 & 0.5659 & 0.4562 & 0.5091 & 0.5714 & \gc\textbf{0.5846} & & 0.5073 & 0.4696 & 0.5624 & 0.5664 & 0.5688 & 0.6365 & \gc\textbf{0.6786} \\
    & LPIPS $\downarrow$ & & 0.7775 & 0.8220 & \textbf{0.6639} & 0.7282 & 0.7085 & 0.8033 & \gc0.7805 & & 0.7480 & 0.7681 & 0.6589 & 0.6604 & 0.6583 & 0.6909 & \gc\textbf{0.6535} \\
    \cmidrule(lr){2-18}
    & PSNR $\uparrow$ & \multirow{4}[0]{*}{\makecell{Gradient \\ Clipping}} & 8.0690 & 7.2342 & 11.3215 & 9.7556 & 9.5759 & 22.2016 & \gc\textbf{23.2074} & \multirow{4}[0]{*}{\makecell{Soteria}} & 6.5675 & 6.6693 & 11.5690 & 9.7588 & 9.0612 & 22.9617 & \gc\textbf{23.7917} \\
    & SSIM $\uparrow$ & & 0.0881 & 0.0659 & 0.2531 & 0.2547 & 0.2390 & 0.6266 & \gc\textbf{0.7023} & & 0.0323 & 0.0382 & 0.2623 & 0.2425 & 0.2166 & 0.6698 & \gc\textbf{0.7143} \\
    & FSIM $\uparrow$ & & 0.5362 & 0.5434 & 0.5900 & 0.5794 & 0.5731 & 0.8229 & \gc\textbf{0.8635} & & 0.4929 & 0.4454 & 0.5964 & 0.5697 & 0.5738 & 0.8430 & \gc\textbf{0.8658} \\
    & LPIPS $\downarrow$ & & 0.7227 & 0.7230 & 0.6117 & 0.6485 & 0.6653 & 0.4603 & \gc\textbf{0.3743} & & 0.7506 & 0.7703 & 0.6031 & 0.6631 & 0.6660 & 0.4253 & \gc\textbf{0.3655} \\
    \bottomrule
    \end{tabular}}
\end{table*}

\begin{table*}[t]
    \caption{Correlation indicators between various training-free search metrics and the actual PSNR performance on CIFAR-10 ($32 \times 32$) and ImageNet ($256 \times 256$) with the default batch size $B = 4$. Note that a correlation indicator closer to 1 (or -1) indicates a stronger positive (or negative) correlation.}
    \label{tab:correlation_indicators}
    \centering
    \tabcolsep=2pt
    \adjustbox{width=\textwidth}{\begin{tabular}{ccccccc}
    \toprule
    \multirow{2}[0]{*}{Training-Free Search Metric} & \multicolumn{3}{c}{CIFAR-10} & \multicolumn{3}{c}{ImageNet} \\
    \cmidrule(lr){2-4} \cmidrule(lr){5-7}
    & Kendall's $\tau$ & Pearson Coefficient & Spearman Coefficient & Kendall's $\tau$ & Pearson Coefficient & Spearman Coefficient \\
    \midrule
    Gaussian Noise $\mathcal{N} \sim (0, 1)$ & 0.0531 & 0.0948 & 0.0720 & 0.1187 & 0.1672 & 0.1773 \\
    Uniform Noise $\mathcal{U} \sim (0, 1)$ & 0.0885 & 0.0921 & 0.1291 & 0.0486 & 0.0822 & 0.0798 \\
    \gc\textbf{Initial Gradient Matching Loss} & \gc\textbf{-0.4491} & \gc\textbf{-0.6128} & \gc\textbf{-0.6345} & \gc\textbf{-0.4906} & \gc\textbf{-0.6633} & \gc\textbf{-0.7120} \\
    \bottomrule
    \end{tabular}}
\end{table*}

\subsection{Further Analysis}
\label{sec:further_analysis}

\begin{figure}[t]
    \centering
    \begin{subfigure}{0.493\linewidth}
        \centering
        \includegraphics[width=\linewidth]{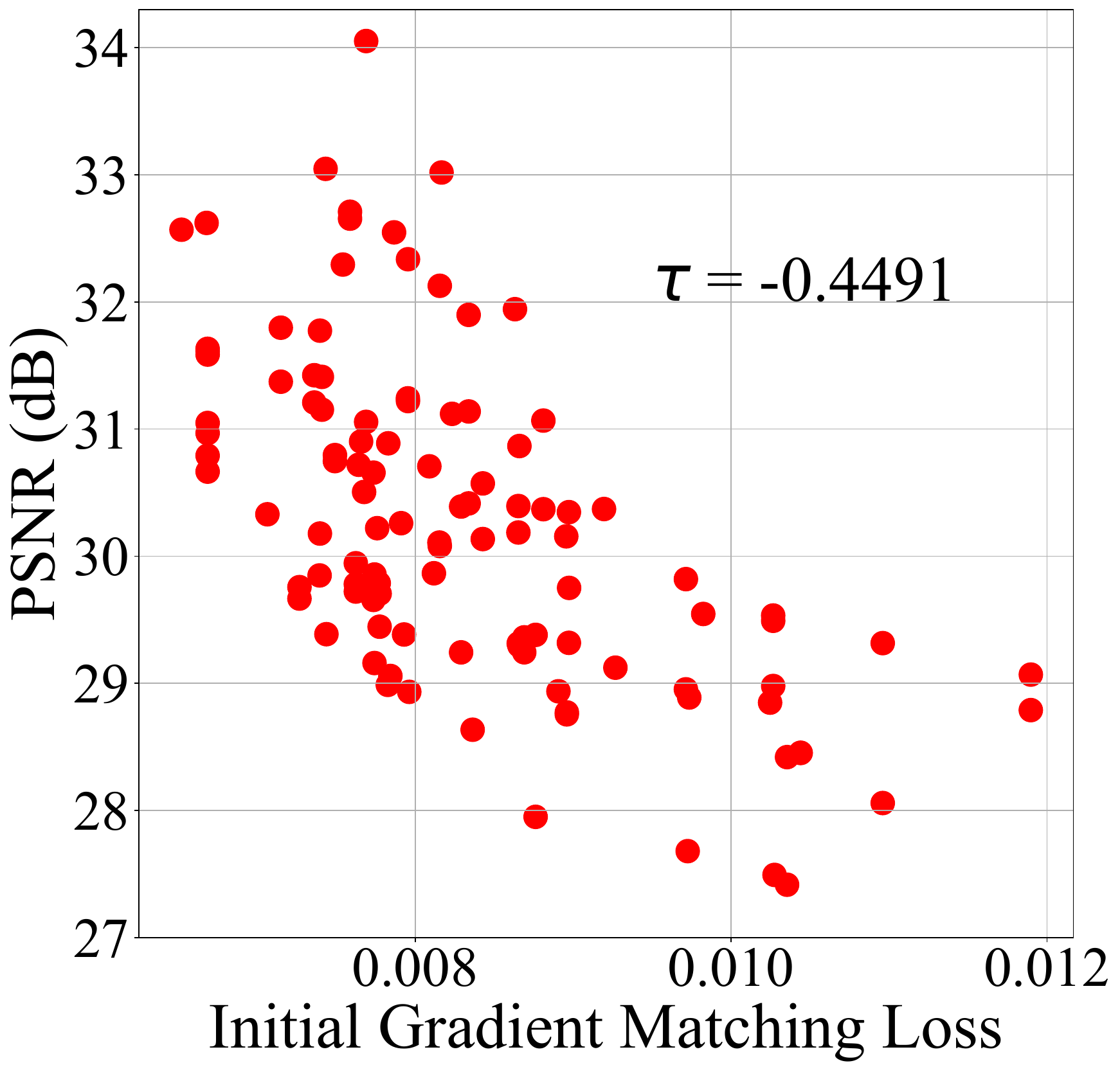}
        \caption{CIFAR-10}
    \end{subfigure}
    \begin{subfigure}{0.493\linewidth}
        \includegraphics[width=\linewidth]{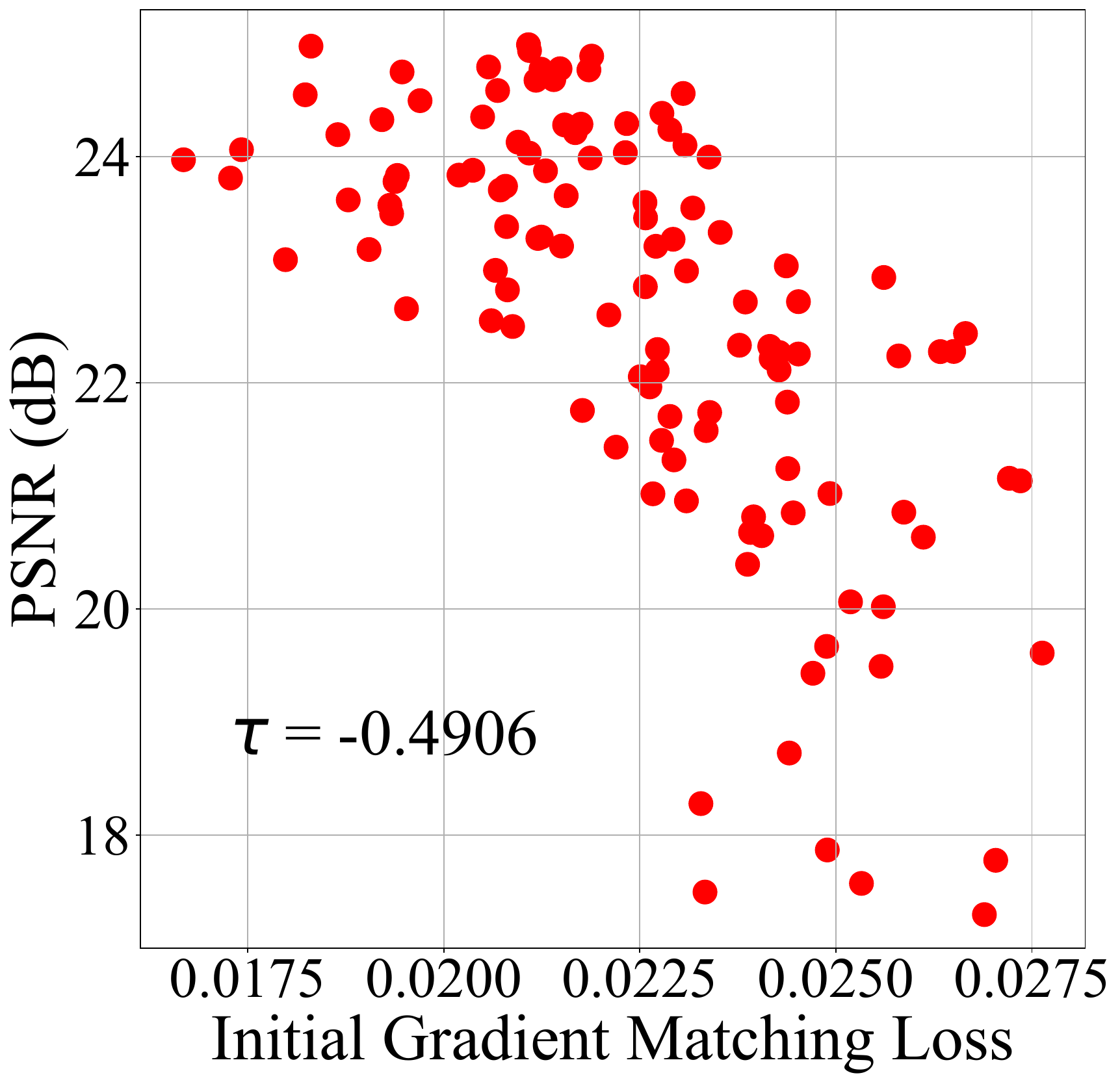}
        \caption{ImageNet}
    \end{subfigure}
    \caption{Correlation between the initial gradient matching loss and the actual PSNR performance on CIFAR-10 ($32 \times 32$) and ImageNet ($256 \times 256$) with the default batch size $B = 4$.}
    \label{correlation_initial_gradient_matching_loss_psnr}
\end{figure}

\textbf{Effectiveness of Our Training-Free Search Metric.} We provide extensive empirical evidence that our training-free search metric (i.e., the initial gradient matching loss) highly negatively correlates with the real attack performance. Specifically, we attack the same private batch using models with different architectures and report the PSNR performance as well as the training-free search metric of each model both on CIFAR-10 and ImageNet. Following previous training-free NAS methods \cite{zhou2022training, zhou2024training, serianni2023training}, we calculate the correlation indicators (i.e., Kendall's $\tau$ \cite{kendall1938new}, Pearson Coefficient \cite{cohen2009pearson}, Spearman Coefficient \cite{spearman1987proof}) to quantitatively measure the relevance. To be more convincing, we introduce Gaussian Noise $\mathcal{N} \sim (0, 1)$ and Uniform Noise $\mathcal{U} \sim (0, 1)$ for comparison, where random noises sampled from $\mathcal{N} \sim (0, 1)$ and $\mathcal{U} \sim (0, 1)$ are regarded as the training-free search metrics. Through this comparative experiment design, we evaluate the performance when using our search method and randomly selecting structures based on Gaussian Noise and Uniform Noise. From Figure \ref{correlation_initial_gradient_matching_loss_psnr} and Table \ref{tab:correlation_indicators}, we conclude that a smaller initial gradient matching loss is more likely to result in better PSNR performance, while the random noises are far from similar effects. Our method significantly outperforms the two random guessing methods based on Gaussian Noise and Uniform Noise. Thus, utilizing the initial gradient matching loss as the training-free search metric is reasonable and meaningful.

\begin{table}[t]
    \caption{Time costs (minute) of different methods averaged over each batch on CIFAR-10 ($32 \times 32$) and ImageNet ($256 \times 256$) with the default batch size $B = 4$.}
    \label{tab:time_costs}
    \centering
    \tabcolsep=2pt
    \adjustbox{width=\linewidth}{\begin{tabular}{ccccccccc}
    \toprule
    Dataset & Metric & IG & GI & GGL & GIAS & GIFD & GION & \gc\textbf{GI-NAS} \\
    \midrule
    \multirow{3}[0]{*}{CIFAR-10} & NAS Time & - & - & - & - & - & - & \gc\textbf{6.6} \\
    & Total Time & 46.1 & 32.9 & 131.4 & 214.8 & \textbf{30.2} & 46.9 & \gc57.3 \\
    & PSNR & 16.3188 & 15.4613 & 12.4938 & 17.3687 & 18.2325 & 30.8652 & \gc\textbf{35.9883} \\
    \midrule
    \multirow{3}[0]{*}{ImageNet} & NAS Time & - & - & - & - & - & - & \gc\textbf{10.9} \\
    & Total Time & 291.1 & 273.7 & 244.6 & 343.2 & 70.6 & \textbf{58.1} & \gc84.4 \\
    & PSNR & 7.9419 & 8.5070 & 11.6255 & 10.0602 & 9.8512 & 21.9942 & \gc\textbf{23.2578} \\
    \bottomrule
    \end{tabular}}
\end{table}

\begin{table}[t]
    \caption{PSNR results of GION and different variants of GI-NAS on CIFAR-10 ($32 \times 32$) and ImageNet ($256 \times 256$) with the default batch size $B = 4$.}
    \label{tab:ablation_study}
    \centering
    \tabcolsep=2pt
    \adjustbox{max width=\linewidth}{\begin{tabular}{ccccc}
    \toprule
    Method & Connection Search & Upsampling Search & CIFAR-10 & ImageNet \\
    \midrule
    GION & \XSolidBrush & \XSolidBrush & 30.8652 & 21.9942 \\
    GI-NAS$^{*}$ & \XSolidBrush & \XSolidBrush & 30.6514 & 22.0126 \\
    GI-NAS$^{\dag}$ & \Checkmark & \XSolidBrush & 35.4336 & 22.4597 \\
    GI-NAS$^{\ddag}$ & \XSolidBrush & \Checkmark & 35.3542 & 22.3992 \\
    \gc\textbf{GI-NAS} & \gc\Checkmark & \gc\Checkmark & \gc\textbf{35.9883} & \gc\textbf{23.2578} \\
    \bottomrule
    \end{tabular}}
\end{table}

\begin{table}[t]
    \caption{Attack performance when increasing the training iterations for the search metric on ImageNet ($256 \times 256$) with the default batch size $B = 4$. Note that the training iteration of $0$ indicates the use of our training-free search metric.}
    \label{tab:ablation_study_training_iteration}
    \centering
    \tabcolsep=2pt
    \adjustbox{max width=\linewidth}{\begin{tabular}{ccccc}
    \toprule
    Search Metric Training Iteration & PSNR $\uparrow$ & SSIM $\uparrow$ & FSIM $\uparrow$ & LPIPS $\downarrow$ \\
    \midrule
    \gc\textbf{0 (Training-Free Search)} & \gc\textbf{23.2578} & \gc\textbf{0.6848} & \gc{0.8513} & \gc\textbf{0.3952} \\
    5 & 22.0149 & 0.6732 & \textbf{0.8545} & 0.3961 \\
    10 & 20.9854 & 0.6115 & 0.8393 & 0.4511 \\
    15 & 21.0370 & 0.6260 & 0.8389 & 0.4358 \\
    \bottomrule
    \end{tabular}}
\end{table}

\begin{table}[t]
    \caption{PSNR results of GI-NAS and state-of-the-art gradient inversion methods when attacking more FL global models (e.g., LeNet-Zhu \cite{zhu2019deep}, ConvNet-32 \cite{geiping2020inverting}) on ImageNet ($256 \times 256$) with the default batch size $B = 4$.}
    \label{tab:more_fl_global_model}
    \centering
    \tabcolsep=2pt
    \adjustbox{width=\linewidth}{\begin{tabular}{cccccccc}
    \toprule
    FL Global Model & IG & GI & GGL & GIAS & GIFD & GION & \gc\textbf{GI-NAS} \\
    \midrule
    LeNet-Zhu \cite{zhu2019deep} & 7.7175 & 6.9530 & 11.8195 & 16.5229 & 17.5620 & 19.2478 & \gc\textbf{21.1847} \\
    ConvNet-32 \cite{geiping2020inverting} & 10.9081 & 10.8886 & 10.4048 & 8.6696 & 24.3937 & 23.9602 & \gc\textbf{25.7824} \\
    ResNet-50 \cite{he2016deep} & 10.9113 & 10.8914 & 9.5354 & 11.0040 & 9.3521 & 9.2080 & \gc\textbf{11.2320} \\
    ViT-Base \cite{dosovitskiy2021an} & 9.7616 & 8.4317 & 9.8555 & 9.9113 & 10.2516 & 11.5427 & \gc\textbf{12.6151} \\
    ViT-Large \cite{dosovitskiy2021an} & 9.6828 & 8.8500 & 9.8445 & 9.9554 & 10.3850 & 11.6536 & \gc\textbf{12.4354} \\
    ViT-Huge \cite{dosovitskiy2021an} & 10.3052 & 9.1382 & 10.5788 & 10.5919 & 10.4953 & 11.7839 & \gc\textbf{13.0380} \\
    \bottomrule
    \end{tabular}}
\end{table}

\textbf{Computational Efficiency.} We next analyze the computational costs. From Table \ref{tab:time_costs}, we find that NAS time only takes around $12\%$ of Total Time, yet GI-NAS still achieves significant PSNR gains over GION and reaches state-of-the-art. The additional time costs introduced by our training-free NAS ($6.6$ or $10.9$ minutes per batch) are very minor, especially given that many methods (e.g., GGL, GIAS) consume over $100$ or even $200$ minutes per batch. Therefore, such minor extra time costs in exchange for significant quantitative improvements are essential and worthwhile. Our method achieves a good trade-off between attack performance and computational efficiency.

\textbf{Ablation Study on Search Type.} In Table \ref{tab:ablation_study}, we report the performance of GION and different variants of GI-NAS. GI-NAS$^{*}$ optimizes on a fixed over-parameterized network with the search size $n = 1$, which means that it is essentially the same as GION. GI-NAS$^{\dag}$ only searches the skip connection patterns while GI-NAS$^{\ddag}$ only searches the upsampling modules. We observe that the performance of GI-NAS$^{*}$ is very close to the performance of GION, as both of them adopt a changeless network architecture. GI-NAS$^{\dag}$ and GI-NAS$^{\ddag}$ improve the attacks on the basis of GION or GI-NAS$^{*}$, which validates the contribution of each search type. GI-NAS combines the above two search types and thus performs the best among all the variants.

\textbf{Ablation Study on Search Metric Training Iteration.} GI-NAS utilizes the initial gradient matching loss as the training-free search metric, which implies that its training iteration for each candidate is $0$. To explore the rationality behind this design, we gradually increase the training iterations for the search metric. From Table \ref{tab:ablation_study_training_iteration}, we conclude that further increasing the training iterations cannot improve the attacks and will instead increase the time costs. Hence, utilizing the initial gradient matching loss to find the optimal models is reasonable and computationally efficient.

\textbf{Generalizability to More FL Global Models.} We show the performance of different methods when attacking more FL global models (e.g., LeNet-Zhu \cite{zhu2019deep}, ConvNet-32 \cite{geiping2020inverting}) in Table \ref{tab:more_fl_global_model}. We note that GI-NAS realizes the best reconstruction results in all the tested cases. These results further demonstrate the reliability of GI-NAS. We also observe significant variations in attack results across different FL global models, even when applying the same gradient inversion method (e.g., a $7.85$ dB PSNR drop when switching from LeNet-Zhu \cite{zhu2019deep} to ConvNet-32 \cite{geiping2020inverting} on GIAS). This implies that future study may have a deeper look into the correlation between the gradient inversion robustness and the architecture of FL global model, and thus design more securing defense strategies from the perspective of adaptive network architecture choice.

\begin{table}[t]
    \caption{Results when attacking the same batch by different networks using \textit{different parameter initialization seeds} on CIFAR-10 ($32 \times 32$) and ImageNet ($256 \times 256$) with the default batch size $B = 4$. \textit{The latent code initialization seeds are fixed for all the models.} $G_{C1} \sim G_{C3}$ and $G_{I1} \sim G_{I3}$ are randomly sampled from the model search space, while $Seed_{C1} \sim Seed_{C3}$ and $Seed_{I1} \sim Seed_{I3}$ are randomly selected and are different from each other. Note that $\sigma_{loss}$ and $\sigma_{psnr}$ are respectively the standard deviations of the initial gradient matching losses and PSNR results on each network.}
    \label{tab:robustness_parameters_initialization}
    \centering
    \tabcolsep=2pt
    \adjustbox{width=\linewidth}{\begin{tabular}{ccccccc}
    \toprule
    Dataset & Model & Random Seed & \makecell{Initial Gradient \\ Matching Loss \\ ($\times 10^{-3}$)} & \makecell{$\sigma_{loss}$ \\ ($\times 10^{-4}$)} & PSNR & $\sigma_{psnr}$ \\
    \midrule
    \multirow{9}[0]{*}{CIFAR-10} & \multirow{3}[0]{*}{$G_{C1}$} & $Seed_{C1}$ & 8.5647 & \multirow{3}[0]{*}{0.0012} & 28.7283 & \multirow{3}[0]{*}{0.1211} \\
    & & $Seed_{C2}$ & 8.5648 & & 28.8082 \\
    & & $Seed_{C3}$ & 8.5645 & & 28.5703 \\
    \cmidrule(lr){2-7}
    & \multirow{3}[0]{*}{$G_{C2}$} & $Seed_{C1}$ & 8.4321 & \multirow{3}[0]{*}{0.0006} & 33.5848 & \multirow{3}[0]{*}{0.2581} \\
    & & $Seed_{C2}$ & 8.4323 & & 33.4631 \\
    & & $Seed_{C3}$ & 8.4322 & & 33.9584 \\
    \cmidrule(lr){2-7}
    & \multirow{3}[0]{*}{$G_{C3}$} & $Seed_{C1}$ & 7.9501 & \multirow{3}[0]{*}{0.0015} & 49.4290 & \multirow{3}[0]{*}{0.2936} \\
    & & $Seed_{C2}$ & 7.9499 & & 49.1171 \\
    & & $Seed_{C3}$ & 7.9502 & & 49.7039 \\
    \midrule
    \multirow{9}[0]{*}{ImageNet} & \multirow{3}[0]{*}{$G_{I1}$} & $Seed_{I1}$ & 21.0167 & \multirow{3}[0]{*}{5.6866} & 22.9013 & \multirow{3}[0]{*}{0.0329} \\
    & & $Seed_{I2}$ & 22.1401 & & 22.8912 \\
    & & $Seed_{I3}$ & 21.4250 & & 22.8398 \\
    \cmidrule(lr){2-7}
    & \multirow{3}[0]{*}{$G_{I2}$} & $Seed_{I1}$ & 24.4600 & \multirow{3}[0]{*}{2.2089} & 21.1475 & \multirow{3}[0]{*}{0.3040} \\
    & & $Seed_{I2}$ & 24.0722 & & 21.6449 \\
    & & $Seed_{I3}$ & 24.0828 & & 21.6989 \\
    \cmidrule(lr){2-7}
    & \multirow{3}[0]{*}{$G_{I3}$} & $Seed_{I1}$ & 18.5319 & \multirow{3}[0]{*}{6.0290} & 23.5278 & \multirow{3}[0]{*}{0.1884} \\
    & & $Seed_{I2}$ & 18.3819 & & 23.2974 \\
    & & $Seed_{I3}$ & 17.4208 & & 23.6709 \\
    \bottomrule
    \end{tabular}}
\end{table}

\textbf{Robustness to Network Parameters Initialization.} Following the general paradigm of previous training-free NAS methods \cite{zhou2022training, zhou2024training, serianni2023training} in other areas, the parameters of all the NAS candidates are randomly initialized and there is no special need to tune the parameters before calculating the initial gradient matching losses. Moreover, GI-NAS records both the architectures as well as the initial parameters when finding the optimal models. Therefore, for the model we decide on, its initial gradient matching loss in the NAS time (i.e., Stage 1 as shown in Figure \ref{fig:pipeline}) is the same as its initial gradient matching loss in the optimization time (i.e., Stage 2 as shown in Figure \ref{fig:pipeline}), and the influences of parameters initialization can be further reduced. To further demonstrate the robustness, we randomly select models with different architectures and apply different parameter initialization seeds to them to attack the same batch. From Table \ref{tab:robustness_parameters_initialization}, we find that for the same architecture, both the initial gradient matching losses and the final PSNR results stay within a steady range when changing the parameter initialization seeds. However, when the architectures change, both the initial gradient matching losses and the final PSNR results will vary significantly. The impacts of different architectures are much greater than the impacts of different parameter initialization seeds. As a result, we conclude that GI-NAS is stable and insensitive to network parameters initialization. We also observe that the models with smaller initial gradient matching losses often have better final PSNR results in Table \ref{tab:robustness_parameters_initialization}, which is consistent with our aforementioned findings in Figure \ref{correlation_initial_gradient_matching_loss_psnr} and Table \ref{tab:correlation_indicators}. This again validates the rationality and usefulness of adopting the initial gradient matching loss as the training-free search metric.

\textbf{Robustness to Latent Codes Initialization.} The latent code $\mathbf{z_0}$ is randomly sampled from the Gaussian distribution $\mathcal{N}(0, 1)$ and then kept frozen. Thus, we explore the influences of different latent codes by changing their initialization seeds. From Table \ref{tab:robustness_latent_codes_initialization}, we discover that although different latent code initialization seeds on the same architecture may result in different initial gradient matching losses, the differences are still very minor (less than $10^{-6}$ for CIFAR-10 and less than $10^{-3}$ for ImageNet), and the impacts on the final PSNR results are limited (less than $0.5$ dB for CIFAR-10 and less than $0.4$ dB for ImageNet). However, when using different architectures, both these two values will change dramatically, and the differences are much greater than those brought by changing the latent code initialization seeds. Hence, GI-NAS is also robust to latent codes initialization.

\begin{table}[t]
    \caption{Results when attacking the same batch by different networks using \textit{different latent code initialization seeds} on CIFAR-10 ($32 \times 32$) and ImageNet ($256 \times 256$) with the default batch size $B = 4$. \textit{The parameter initialization seeds are fixed for all the models.} $G_{C4} \sim G_{C6}$ and $G_{I4} \sim G_{I6}$ are randomly sampled from the model search space, while $Seed_{C4} \sim Seed_{C6}$ and $Seed_{I4} \sim Seed_{I6}$ are randomly selected and are different from each other. Note that $\sigma_{loss}$ and $\sigma_{psnr}$ are respectively the standard deviations of the initial gradient matching losses and PSNR results on each network.}
    \label{tab:robustness_latent_codes_initialization}
    \centering
    \tabcolsep=2pt
    \adjustbox{width=\linewidth}{\begin{tabular}{ccccccc}
    \toprule
    Dataset & Model & Random Seed & \makecell{Initial Gradient \\ Matching Loss \\ ($\times 10^{-3}$)} & \makecell{$\sigma_{loss}$ \\ ($\times 10^{-4}$)} & PSNR & $\sigma_{psnr}$ \\
    \midrule
    \multirow{9}[0]{*}{CIFAR-10} & \multirow{3}[0]{*}{$G_{C4}$} & $Seed_{C4}$ & 14.5338 & \multirow{3}[0]{*}{0.0018} & 26.4506 & \multirow{3}[0]{*}{0.1925} \\
    & & $Seed_{C5}$ & 14.5335 & & 26.2082 \\
    & & $Seed_{C6}$ & 14.5336 & & 26.0703 \\
    \cmidrule(lr){2-7}
    & \multirow{3}[0]{*}{$G_{C5}$} & $Seed_{C4}$ & 13.4751 & \multirow{3}[0]{*}{0.0018} & 30.1240 & \multirow{3}[0]{*}{0.2675} \\
    & & $Seed_{C5}$ & 13.4753 & & 30.2045 \\
    & & $Seed_{C6}$ & 13.4755 & & 30.6222 \\
    \cmidrule(lr){2-7}
    & \multirow{3}[0]{*}{$G_{C6}$} & $Seed_{C4}$ & 13.2704 & \multirow{3}[0]{*}{0.0015} & 31.3139 & \multirow{3}[0]{*}{0.0566} \\
    & & $Seed_{C5}$ & 13.2701 & & 31.4256 \\
    & & $Seed_{C6}$ & 13.2702 & & 31.3540 \\
    \midrule
    \multirow{9}[0]{*}{ImageNet} & \multirow{3}[0]{*}{$G_{I4}$} & $Seed_{I4}$ & 23.1931 & \multirow{3}[0]{*}{4.6280} & 21.8376 & \multirow{3}[0]{*}{0.1164} \\
    & & $Seed_{I5}$ & 23.6335 & & 22.0556 \\
    & & $Seed_{I6}$ & 22.7083 & & 21.8760 \\
    \cmidrule(lr){2-7}
    & \multirow{3}[0]{*}{$G_{I5}$} & $Seed_{I4}$ & 19.5264 & \multirow{3}[0]{*}{1.1102} & 23.6033 & \multirow{3}[0]{*}{0.1889} \\
    & & $Seed_{I5}$ & 19.5677 & & 23.2439 \\
    & & $Seed_{I6}$ & 19.7360 & & 23.3226 \\
    \cmidrule(lr){2-7}
    & \multirow{3}[0]{*}{$G_{I6}$} & $Seed_{I4}$ & 25.9479 & \multirow{3}[0]{*}{3.6054} & 20.4730 & \multirow{3}[0]{*}{0.1655} \\
    & & $Seed_{I5}$ & 25.3497 & & 20.8007 \\
    & & $Seed_{I6}$ & 25.9975 & & 20.6776 \\
    \bottomrule
    \end{tabular}}
\end{table}

\begin{table*}[t]
    \caption{Averaged architectural statistics of the searched optimal models by our NAS strategy under various settings tested in Section \ref{sec:main_comparison}. The default settings are highlighted in gray color. $\Delta_{connection}$ and $\Delta_{upsampling}$ are respectively the gains of the number of skip connections and the kernel size of upsampling modules when compared with the default settings.}
    \label{tab:nas_preferences}
    \centering
    \tabcolsep=2pt
    \adjustbox{width=\linewidth}{\begin{tabular}{ccccccccc}
    \toprule
    Dataset & Batch Size & Defense & \makecell{Number of \\ Skip Connections} & \makecell{Kernel Size of \\ Upsampling Modules} & $\Delta_{connection}$ & $\Delta_{upsampling}$ & \makecell{Mostly Used \\ Upsampling Operations} \\
    \midrule
    \multirow{9}[0]{*}{CIFAR-10} & \gc4 & \gc{No Defense} & \gc11.4 & \gc4.8 & \gc0.0 & \gc0.0 & \gc{bilinear} \\
    & 16 & No Defense & 13.5 & 4.0 & 2.1 & -0.8 & bilinear \\
    & 32 & No Defense & 14.0 & 5.0 & 2.6 & 0.3 & bicubic \\
    & 48 & No Defense & 13.0 & 5.0 & 1.6 & 0.3 & bicubic \\
    & 96 & No Defense & 12.0 & 7.0 & 0.6 & 2.3 & bicubic \\
    \cmidrule{2-8}
    & 4 & Gaussian Noise & 12.3 & 5.5 & 0.9 & 0.8 & bilinear \\
    & 4 & Gradient Sparsification & 11.8 & 4.5 & 0.4 & -0.3 & pixel shuffle \\
    & 4 & Gaussian Clipping & 12.5 & 5.0 & 1.1 & 0.3 & bilinear \\
    & 4 & Soteria & 11.5 & 5.0 & 0.1 & 0.3 & bicubic \\
    \midrule
    \multirow{9}[0]{*}{ImageNet} & \gc4 & \gc{No Defense} & \gc11.9 & \gc3.8 & \gc0.0 & \gc0.0 & \gc{bilinear} \\
    & 8 & No Defense & 12.8 & 5.5 & 0.9 & 1.8 & bilinear \\
    & 16 & No Defense & 14.0 & 6.0 & 2.1 & 2.3 & bilinear \\
    & 24 & No Defense & 16.0 & 5.0 & 4.1 & 1.3 & bilinear \\
    & 32 & No Defense & 12.0 & 7.0 & 0.1 & 3.3 & bicubic \\
    \cmidrule{2-8}
    & 4 & Gaussian Noise & 12.3 & 5.0 & 0.4 & 1.3 & pixel shuffle \\
    & 4 & Gradient Sparsification & 12.4 & 4.3 & 0.5 & 0.5 & bilinear \\
    & 4 & Gaussian Clipping & 14.1 & 3.3 & 2.3 & -0.5 & bilinear \\
    & 4 & Soteria & 12.6 & 4.3 & 0.8 & 0.5 & bilinear \\
    \bottomrule
    \end{tabular}}
\end{table*}

\textbf{Preferences of NAS Outcomes.} To figure out the implications behind NAS preferences, we trace back the searched optimal models from Section \ref{sec:main_comparison} and analyze their averaged architectural statistics in Table \ref{tab:nas_preferences}. Note that the bicubic operation takes more neighbourhood pixels than the bilinear operation ($16 > 4$) and is thus considered to be more complex. Besides, the pixel shuffle operation rearranges lots of pixels and is also considered to be more complicated than the bilinear operation. From Table \ref{tab:nas_preferences}, we observe that the optimal models are largely dependent on recovery difficulties. Generally, batches with larger sizes or under defenses are more difficult to recover. Compared to the default settings (highlighted in gray color) that are with the batch size $B = 4$ and under no defenses, the more difficult settings would prefer models with higher complexity (e.g., more skip connections, larger kernel sizes, more complex operations). The complexity gains $\Delta_{connection}$ and $\Delta_{upsampling}$ are often greater than $0$ when the settings are $B > 4$ or under defenses. Easier batches prefer simpler, lighter architectures and harder batches prefer more complex, heavier architectures. This might be used to further improve the efficiency. By considering the difficulties ahead of time, we can have a better estimation of the complexity of optimal models. Thus, we only need to focus on models with a certain type of complexity and skip other unrelated models even when there is not enough time for NAS.
\section{Conclusion}

In this paper, we propose GI-NAS, a novel gradient inversion method that makes deeper use of the implicit architectural priors for gradient inversion. We first systematically analyze existing gradient inversion methods and emphasize the necessity of adaptive architecture search. We then build up our model search space by designing different upsampling modules and skip connection patterns. To reduce the computational overhead, we leverage the initial gradient matching loss as the training-free search metric to select the optimal model architecture and provide extensive experimental evidence that such a metric highly correlates with the real attack performance. Extensive experiments show that GI-NAS can achieve state-of-the-art performance compared to existing methods, even under more practical FL scenarios with high-resolution images, large-sized batches, and advanced defense strategies. We also provide deeper analysis on various aspects, such as time costs, ablation studies, generalizability to more FL global models, robustness to network parameters initialization, robustness to latent codes initialization, and implications behind the NAS searched results. We hope that the remarkable attack improvements of GI-NAS over existing gradient inversion methods may help raise the public awareness of such privacy threats, as the sensitive data would be more likely to be revealed or even abused. Moreover, we hope that the idea of this paper may shed new light on the gradient inversion community and facilitate the research in this field.

\textbf{Limitations and Future Directions.} Although we have empirically demonstrated that the initial gradient matching loss is highly negatively correlated to the final attack performance, it is still of great significance to further prove the effectiveness of this search metric with rigorous theoretical analysis. In the future work, we hope to provide more insightful theoretical results with regard to this search metric from various perspectives, such as the frequency spectrum \cite{liu2023devil, arican2022isnas} or the implicit neural architectural priors \cite{ulyanov2018deep, cheng2019bayesian}. Furthermore, inspired by the effectiveness of NAS for attacks in this paper, a promising defensive direction involves exploring victim-side neural architecture search to discover model structures inherently more resistant to gradient inversion while maintaining model accuracy, or developing adaptive gradient perturbation techniques on the FL global model guided by NAS principles.

\bibliographystyle{IEEEtran}
\bibliography{references}

\appendices{}

\section{Attack Results on More Datasets}

We further compare the attack effectiveness of GI-NAS with other methods on more datasets, including FFHQ \cite{karras2019style}, MNIST \cite{deng2012mnist}, and SVHN \cite{netzer2011reading}. These datasets are also considered in previous gradient inversion methods \cite{liang2023egia, fang2023gifd}. From Table \ref{tab:results_more_datasets}, we observe that GI-NAS consistently outperforms all the compared methods across diverse datasets. This consistent superiority suggests that GI-NAS is a highly generalizable and effective method for privacy attacks in federated learning.

\begin{table}[h]
    \caption{Quantitative comparison of GI-NAS to state-of-the-art gradient inversion methods on FFHQ ($64 \times 64$), MNIST ($28 \times 28$), and SVHN ($32 \times 32$) with the default batch size $B = 4$.}
    \label{tab:results_more_datasets}
    \centering
    \tabcolsep=2pt
    \adjustbox{max width=\linewidth}{\begin{tabular}{ccccccccc}
    \toprule
    Dataset & Metric & IG & GI & GGL & GIAS & GIFD & GION & \gc\textbf{GI-NAS} \\
    \midrule
    \multirow{4}[0]{*}{FFHQ} & PSNR $\uparrow$ & 12.1571 & 12.4204 & 10.8660 & 13.2693 & 13.7942 & 31.7174 & \gc\textbf{36.2522} \\
    & SSIM $\uparrow$ & 0.2226 & 0.2237 & 0.1958 & 0.3418 & 0.3609 & 0.9865 & \gc\textbf{0.9954} \\
    & FSIM $\uparrow$ & 0.6212 & 0.6128 & 0.6099 & 0.6623 & 0.6763 & 0.9879 & \gc\textbf{0.9979} \\
    & LPIPS $\downarrow$ & 0.7448 & 0.7516 & 0.6327 & 0.6160 & 0.6182 & 0.0212 & \gc\textbf{0.0025} \\
    \midrule
    \multirow{4}[0]{*}{MNIST} & PSNR $\uparrow$ & 15.1129 & 14.1863 & 9.3894 & 22.8035 & 28.8421 & 31.9038 & \gc\textbf{37.4773} \\
    & SSIM $\uparrow$ & 0.3497 & 0.3220 & 0.0670 & 0.5962 & 0.7874 & 0.9935 & \gc\textbf{0.9953} \\
    & FSIM $\uparrow$ & 0.5319 & 0.5115 & 0.3957 & 0.7520 & 0.8959 & 0.9982 & \gc\textbf{0.9993} \\
    & LPIPS $\downarrow$ & 0.3898 & 0.4425 & 0.5735 & 0.2240 & 0.0923 & 0.0014 & \gc\textbf{0.0005} \\
    \midrule
    \multirow{4}[0]{*}{SVHN} & PSNR $\uparrow$ & 20.9779 & 20.9475 & 16.9157 & 21.0760 & 22.8974 & 44.8053 & \gc\textbf{46.4780} \\
    & SSIM $\uparrow$ & 0.6702 & 0.7047 & 0.4427 & 0.6704 & 0.7894 & 0.9774 & \gc\textbf{0.9870} \\
    & FSIM $\uparrow$ & 0.7993 & 0.8121 & 0.6535 & 0.8045 & 0.8531 & 0.9895 & \gc\textbf{0.9941} \\
    & LPIPS $\downarrow$ & 0.5145 & 0.5247 & 0.6789 & 0.5148 & 0.4270 & 0.0045 & \gc\textbf{0.0028} \\
    \bottomrule
    \end{tabular}}
\end{table}

\section{Attack Results on Heterogeneous Data}

FL systems would commonly face the problem of heterogeneous data \cite{ye2023heterogeneous}. Thus, to further validate the robustness of our method, we strictly align with the prior works \cite{wei2025extracting, cao2021fltrust} to design the data heterogeneity experiments. To be specific, we randomly allocate the classes of datasets into $N_{client}$ clients with a probability $q$ of assigning a training example with a certain label to a particular client and a probability $\frac{1 - q}{N_{client} - 1}$ of assigning it to other clients. When $q = \frac{1}{N_{client}}$, the client's local training data are independent and identically distributed (IID), otherwise the client's local training data are Non-IID.

\begin{figure}[h]
    \centering
    \includegraphics[width=\linewidth]{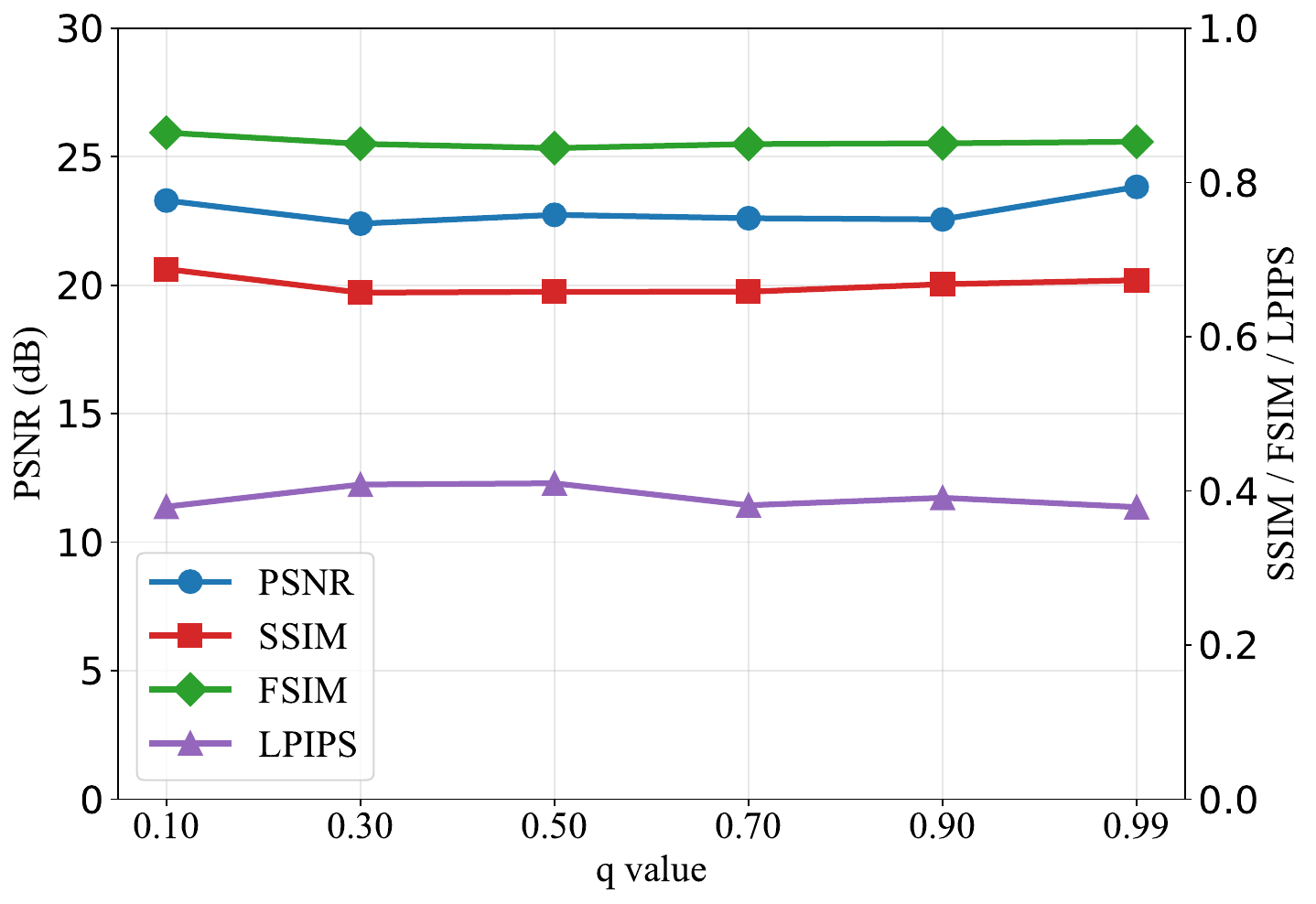}
    \caption{Attack performance of GI-NAS when changing the values of $q$ (the probability of assigning a training example with a certain label to a particular client) on ImageNet ($256 \times 256$) with the default batch size $B = 4$. Note that there are $N_{client} = 10$ clients in total, which indicates that if $q = \frac{1}{10}$, the distribution is IID. These settings strictly adhere to the settings in previous works \cite{wei2025extracting, cao2021fltrust} considering data heterogeneity.}
    \label{fig:data_heterogeneity}
\end{figure}

We set $N_{client} = 10$, which means that if $q = \frac{1}{10}$, the distribution is IID. We select $q = 0.10, 0.30, 0.50, 0.70, 0.90, 0.99$, and test the performance of GI-NAS under these settings. From Figure \ref{fig:data_heterogeneity}, we notice that the attack performance of GI-NAS stays within a steady range when changing the values of $q$. Therefore, we conclude that GI-NAS is also robust to the influences of data heterogeneity.

\section{More Visualized Results}

\begin{figure*}[b]

        \begin{subfigure}{\linewidth}
        \centering
        \begin{minipage}[t]{0.1195\linewidth}
        \centering
        \includegraphics[width=2.245cm]{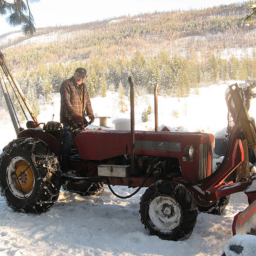}
        \centering
        \end{minipage}
        \begin{minipage}[t]{0.1195\linewidth}
        \centering
        \includegraphics[width=2.245cm]{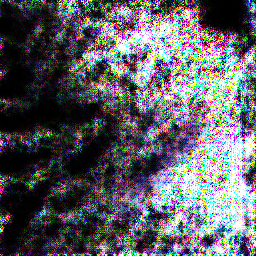}
        \centering
        \end{minipage}
        \begin{minipage}[t]{0.1195\linewidth}
        \centering
        \includegraphics[width=2.245cm]{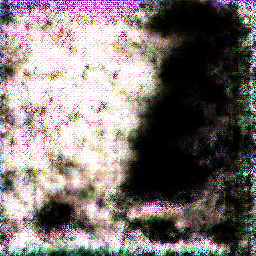}
        \centering
        \end{minipage}
        \begin{minipage}[t]{0.1195\linewidth}
        \centering
        \includegraphics[width=2.245cm]{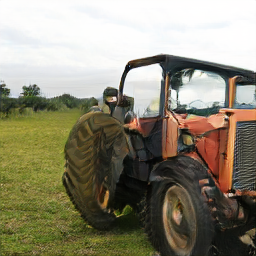}
        \centering
        \end{minipage}
        \begin{minipage}[t]{0.1195\linewidth}
        \centering
        \includegraphics[width=2.245cm]{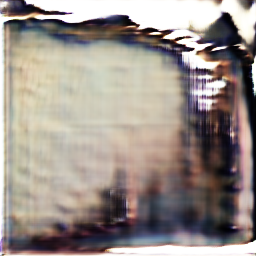}
        \centering
        \end{minipage}
        \begin{minipage}[t]{0.1195\linewidth}
        \centering
        \includegraphics[width=2.245cm]{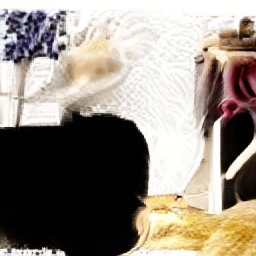}
        \centering
        \end{minipage}
        \begin{minipage}[t]{0.1195\linewidth}
        \centering
        \includegraphics[width=2.245cm]{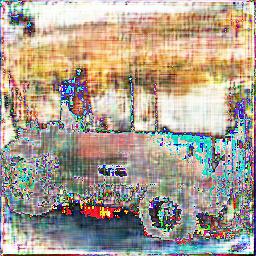}
        \centering
        \end{minipage}
        \begin{minipage}[t]{0.1195\linewidth}
        \centering
        \includegraphics[width=2.245cm]{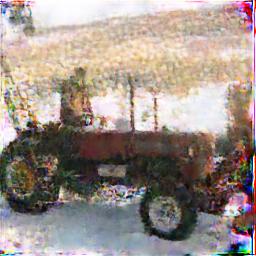}
        \centering
        \end{minipage}
        \end{subfigure}

        \begin{subfigure}{\linewidth}
        \centering
        \begin{minipage}[t]{0.1195\linewidth}
        \centering
        \includegraphics[width=2.245cm]{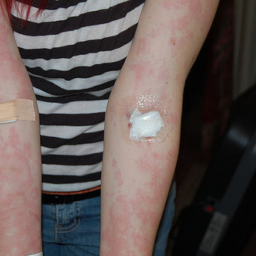}
        \centering
        \end{minipage}
        \begin{minipage}[t]{0.1195\linewidth}
        \centering
        \includegraphics[width=2.245cm]{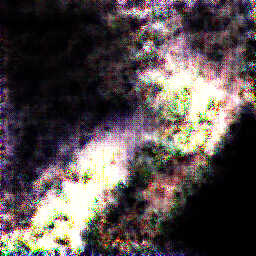}
        \centering
        \end{minipage}
        \begin{minipage}[t]{0.1195\linewidth}
        \centering
        \includegraphics[width=2.245cm]{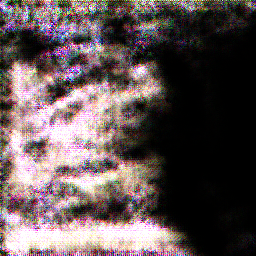}
        \centering
        \end{minipage}
        \begin{minipage}[t]{0.1195\linewidth}
        \centering
        \includegraphics[width=2.245cm]{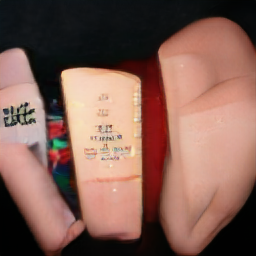}
        \centering
        \end{minipage}
        \begin{minipage}[t]{0.1195\linewidth}
        \centering
        \includegraphics[width=2.245cm]{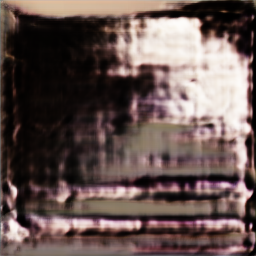}
        \centering
        \end{minipage}
        \begin{minipage}[t]{0.1195\linewidth}
        \centering
        \includegraphics[width=2.245cm]{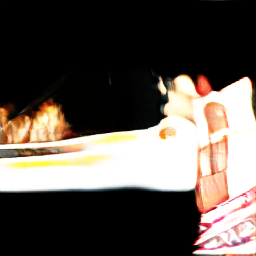}
        \centering
        \end{minipage}
        \begin{minipage}[t]{0.1195\linewidth}
        \centering
        \includegraphics[width=2.245cm]{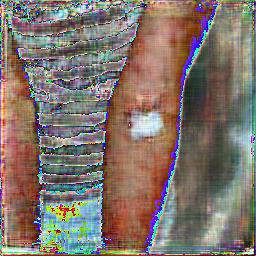}
        \centering
        \end{minipage}
        \begin{minipage}[t]{0.1195\linewidth}
        \centering
        \includegraphics[width=2.245cm]{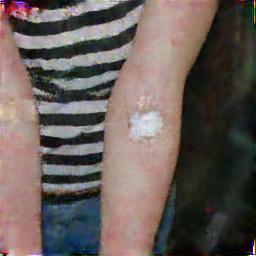}
        \centering
        \end{minipage}
        \end{subfigure}

        \begin{subfigure}{\linewidth}
        \centering
        \begin{minipage}[t]{0.1195\linewidth}
        \centering
        \includegraphics[width=2.245cm]{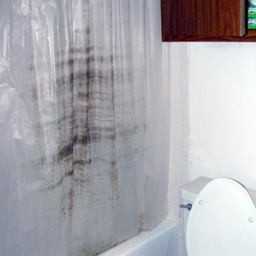}
        \centering
        \end{minipage}
        \begin{minipage}[t]{0.1195\linewidth}
        \centering
        \includegraphics[width=2.245cm]{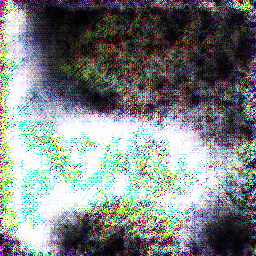}
        \centering
        \end{minipage}
        \begin{minipage}[t]{0.1195\linewidth}
        \centering
        \includegraphics[width=2.245cm]{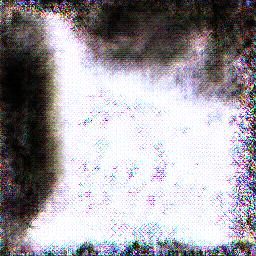}
        \centering
        \end{minipage}
        \begin{minipage}[t]{0.1195\linewidth}
        \centering
        \includegraphics[width=2.245cm]{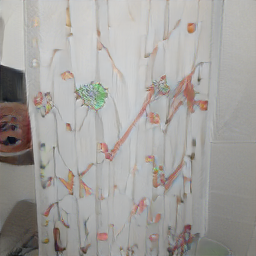}
        \centering
        \end{minipage}
        \begin{minipage}[t]{0.1195\linewidth}
        \centering
        \includegraphics[width=2.245cm]{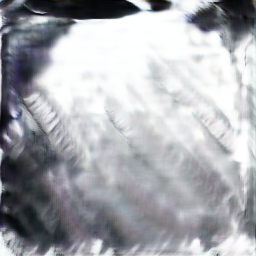}
        \centering
        \end{minipage}
        \begin{minipage}[t]{0.1195\linewidth}
        \centering
        \includegraphics[width=2.245cm]{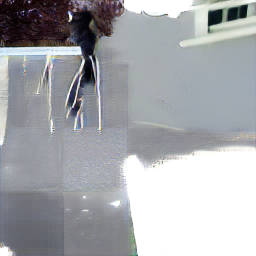}
        \centering
        \end{minipage}
        \begin{minipage}[t]{0.1195\linewidth}
        \centering
        \includegraphics[width=2.245cm]{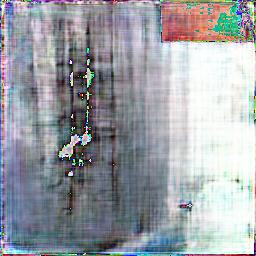}
        \centering
        \end{minipage}
        \begin{minipage}[t]{0.1195\linewidth}
        \centering
        \includegraphics[width=2.245cm]{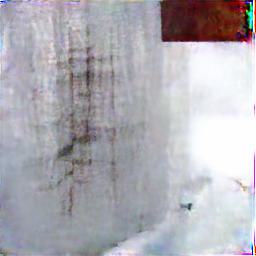}
        \centering
        \end{minipage}
        \end{subfigure}

        \begin{subfigure}{\linewidth}
        \centering
        \begin{minipage}[t]{0.1195\linewidth}
        \centering
        \includegraphics[width=2.245cm]{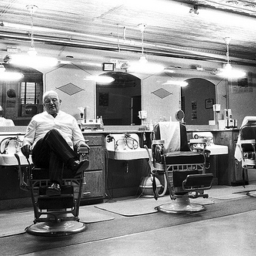}
        \centering
        \end{minipage}
        \begin{minipage}[t]{0.1195\linewidth}
        \centering
        \includegraphics[width=2.245cm]{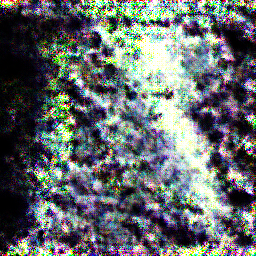}
        \centering
        \end{minipage}
        \begin{minipage}[t]{0.1195\linewidth}
        \centering
        \includegraphics[width=2.245cm]{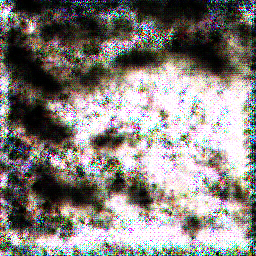}
        \centering
        \end{minipage}
        \begin{minipage}[t]{0.1195\linewidth}
        \centering
        \includegraphics[width=2.245cm]{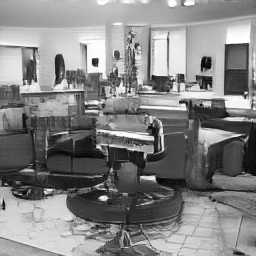}
        \centering
        \end{minipage}
        \begin{minipage}[t]{0.1195\linewidth}
        \centering
        \includegraphics[width=2.245cm]{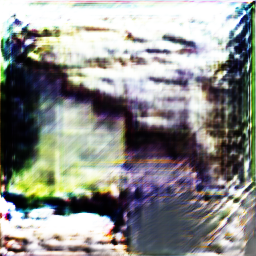}
        \centering
        \end{minipage}
        \begin{minipage}[t]{0.1195\linewidth}
        \centering
        \includegraphics[width=2.245cm]{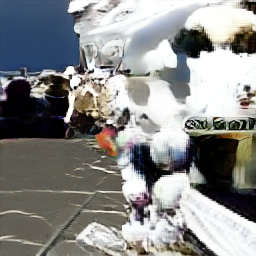}
        \centering
        \end{minipage}
        \begin{minipage}[t]{0.1195\linewidth}
        \centering
        \includegraphics[width=2.245cm]{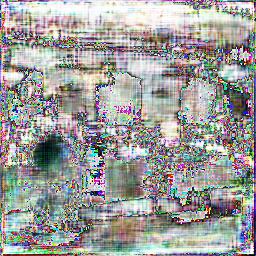}
        \centering
        \end{minipage}
        \begin{minipage}[t]{0.1195\linewidth}
        \centering
        \includegraphics[width=2.245cm]{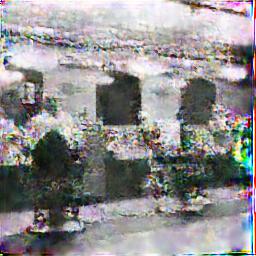}
        \centering
        \end{minipage}
        \end{subfigure}

        \begin{subfigure}{\linewidth}
        \centering
        \begin{minipage}[t]{0.1195\linewidth}
        \centering
        \includegraphics[width=2.245cm]{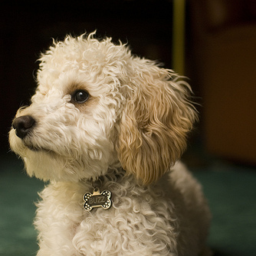}
        \centering
        \end{minipage}
        \begin{minipage}[t]{0.1195\linewidth}
        \centering
        \includegraphics[width=2.245cm]{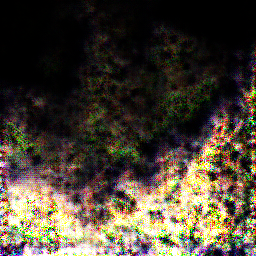}
        \centering
        \end{minipage}
        \begin{minipage}[t]{0.1195\linewidth}
        \centering
        \includegraphics[width=2.245cm]{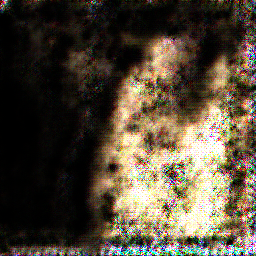}
        \centering
        \end{minipage}
        \begin{minipage}[t]{0.1195\linewidth}
        \centering
        \includegraphics[width=2.245cm]{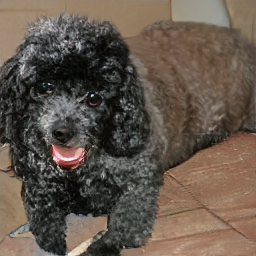}
        \centering
        \end{minipage}
        \begin{minipage}[t]{0.1195\linewidth}
        \centering
        \includegraphics[width=2.245cm]{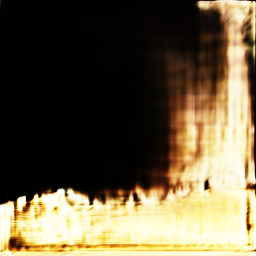}
        \centering
        \end{minipage}
        \begin{minipage}[t]{0.1195\linewidth}
        \centering
        \includegraphics[width=2.245cm]{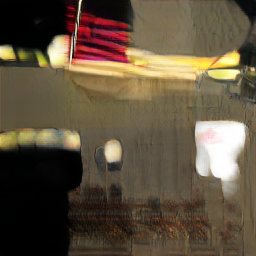}
        \centering
        \end{minipage}
        \begin{minipage}[t]{0.1195\linewidth}
        \centering
        \includegraphics[width=2.245cm]{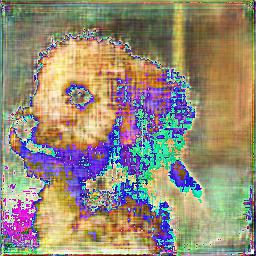}
        \centering
        \end{minipage}
        \begin{minipage}[t]{0.1195\linewidth}
        \centering
        \includegraphics[width=2.245cm]{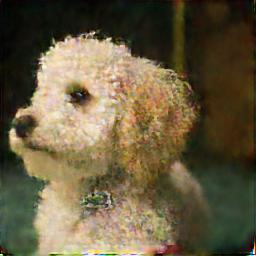}
        \centering
        \end{minipage}
        \end{subfigure}

        \begin{subfigure}{\linewidth}
        \centering
        \begin{minipage}[t]{0.1195\linewidth}
        \centering
        \includegraphics[width=2.245cm]{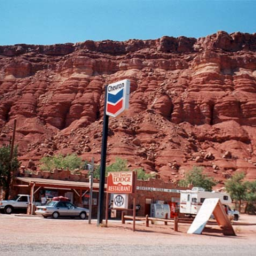}
        \centering
        \end{minipage}
        \begin{minipage}[t]{0.1195\linewidth}
        \centering
        \includegraphics[width=2.245cm]{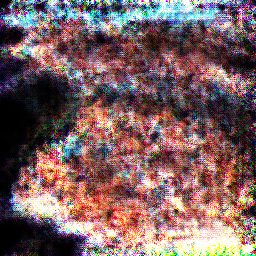}
        \centering
        \end{minipage}
        \begin{minipage}[t]{0.1195\linewidth}
        \centering
        \includegraphics[width=2.245cm]{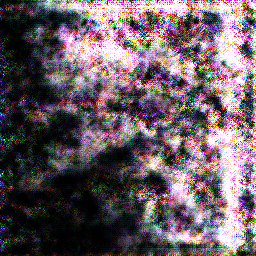}
        \centering
        \end{minipage}
        \begin{minipage}[t]{0.1195\linewidth}
        \centering
        \includegraphics[width=2.245cm]{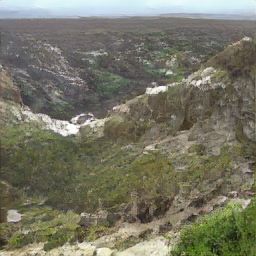}
        \centering
        \end{minipage}
        \begin{minipage}[t]{0.1195\linewidth}
        \centering
        \includegraphics[width=2.245cm]{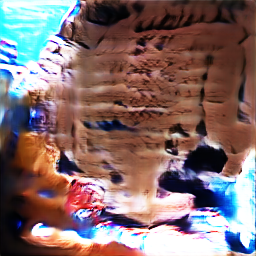}
        \centering
        \end{minipage}
        \begin{minipage}[t]{0.1195\linewidth}
        \centering
        \includegraphics[width=2.245cm]{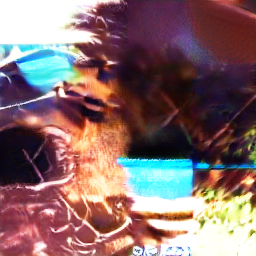}
        \centering
        \end{minipage}
        \begin{minipage}[t]{0.1195\linewidth}
        \centering
        \includegraphics[width=2.245cm]{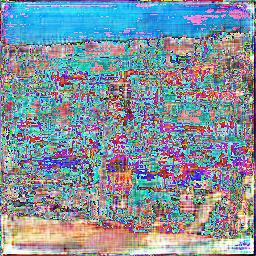}
        \centering
        \end{minipage}
        \begin{minipage}[t]{0.1195\linewidth}
        \centering
        \includegraphics[width=2.245cm]{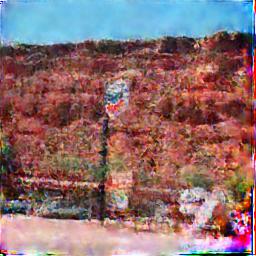}
        \centering
        \end{minipage}
        \end{subfigure}

        \begin{subfigure}{\linewidth}
        \centering
        \begin{minipage}[t]{0.1195\linewidth}
        \centering
        \includegraphics[width=2.245cm]{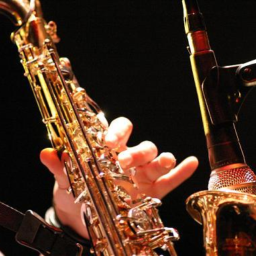}
        \centering
        \end{minipage}
        \begin{minipage}[t]{0.1195\linewidth}
        \centering
        \includegraphics[width=2.245cm]{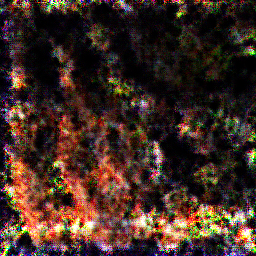}
        \centering
        \end{minipage}
        \begin{minipage}[t]{0.1195\linewidth}
        \centering
        \includegraphics[width=2.245cm]{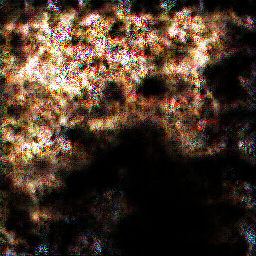}
        \centering
        \end{minipage}
        \begin{minipage}[t]{0.1195\linewidth}
        \centering
        \includegraphics[width=2.245cm]{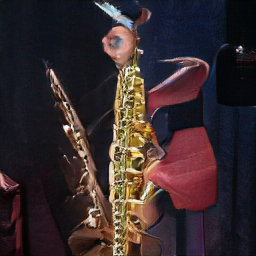}
        \centering
        \end{minipage}
        \begin{minipage}[t]{0.1195\linewidth}
        \centering
        \includegraphics[width=2.245cm]{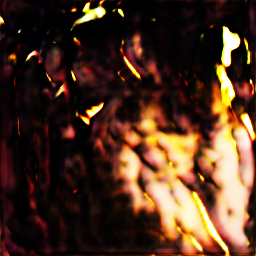}
        \centering
        \end{minipage}
        \begin{minipage}[t]{0.1195\linewidth}
        \centering
        \includegraphics[width=2.245cm]{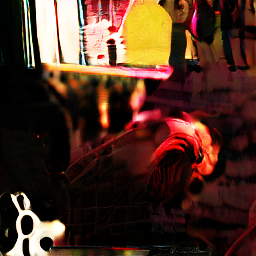}
        \centering
        \end{minipage}
        \begin{minipage}[t]{0.1195\linewidth}
        \centering
        \includegraphics[width=2.245cm]{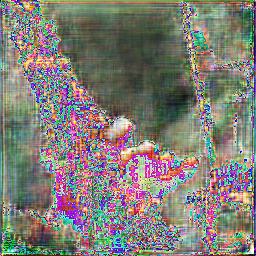}
        \centering
        \end{minipage}
        \begin{minipage}[t]{0.1195\linewidth}
        \centering
        \includegraphics[width=2.245cm]{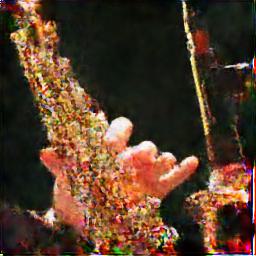}
        \centering
        \end{minipage}
        \end{subfigure}

        \begin{subfigure}{\linewidth}
        \centering
        \begin{minipage}[t]{0.1195\linewidth}
        \centering
        \includegraphics[width=2.245cm]{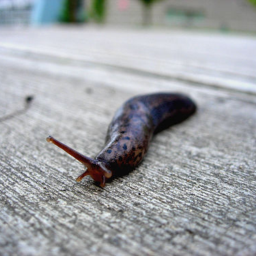}
        \centering
        \end{minipage}
        \begin{minipage}[t]{0.1195\linewidth}
        \centering
        \includegraphics[width=2.245cm]{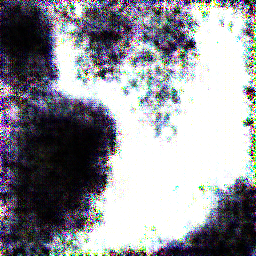}
        \centering
        \end{minipage}
        \begin{minipage}[t]{0.1195\linewidth}
        \centering
        \includegraphics[width=2.245cm]{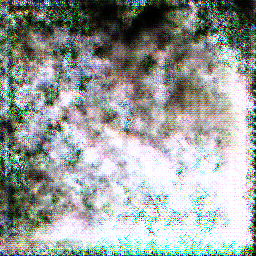}
        \centering
        \end{minipage}
        \begin{minipage}[t]{0.1195\linewidth}
        \centering
        \includegraphics[width=2.245cm]{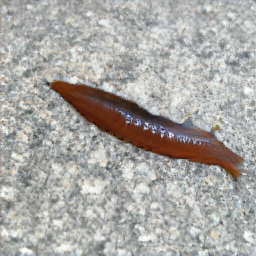}
        \centering
        \end{minipage}
        \begin{minipage}[t]{0.1195\linewidth}
        \centering
        \includegraphics[width=2.245cm]{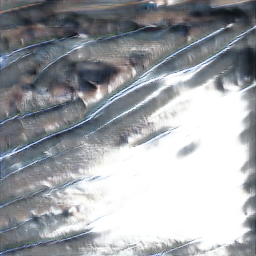}
        \centering
        \end{minipage}
        \begin{minipage}[t]{0.1195\linewidth}
        \centering
        \includegraphics[width=2.245cm]{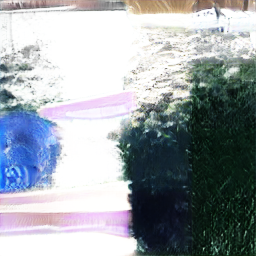}
        \centering
        \end{minipage}
        \begin{minipage}[t]{0.1195\linewidth}
        \centering
        \includegraphics[width=2.245cm]{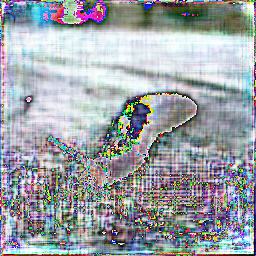}
        \centering
        \end{minipage}
        \begin{minipage}[t]{0.1195\linewidth}
        \centering
        \includegraphics[width=2.245cm]{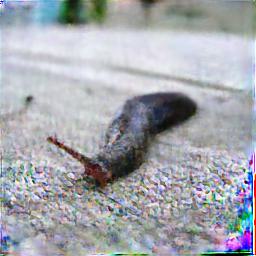}
        \centering
        \end{minipage}
        \end{subfigure}

        \vspace{0.9pt}

        \begin{subfigure}{\linewidth}
        \centering
        \begin{minipage}[t]{0.1195\linewidth}
        \centering
        \includegraphics[width=2.245cm]{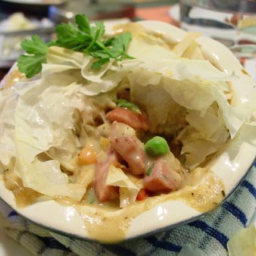}
        \centering
        \end{minipage}
        \begin{minipage}[t]{0.1195\linewidth}
        \centering
        \includegraphics[width=2.245cm]{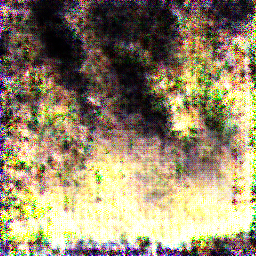}
        \centering
        \end{minipage}
        \begin{minipage}[t]{0.1195\linewidth}
        \centering
        \includegraphics[width=2.245cm]{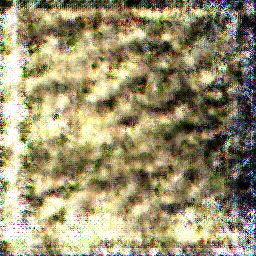}
        \centering
        \end{minipage}
        \begin{minipage}[t]{0.1195\linewidth}
        \centering
        \includegraphics[width=2.245cm]{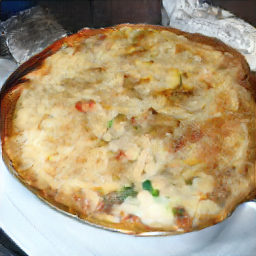}
        \centering
        \end{minipage}
        \begin{minipage}[t]{0.1195\linewidth}
        \centering
        \includegraphics[width=2.245cm]{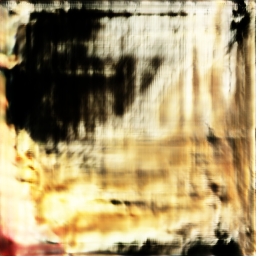}
        \centering
        \end{minipage}
        \begin{minipage}[t]{0.1195\linewidth}
        \centering
        \includegraphics[width=2.245cm]{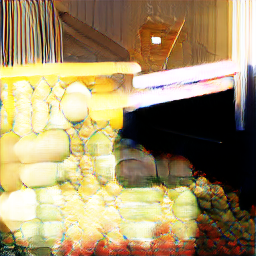}
        \centering
        \end{minipage}
        \begin{minipage}[t]{0.1195\linewidth}
        \centering
        \includegraphics[width=2.245cm]{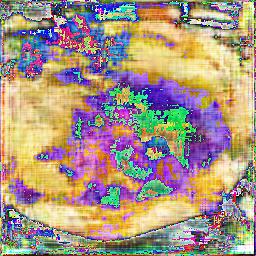}
        \centering
        \end{minipage}
        \begin{minipage}[t]{0.1195\linewidth}
        \centering
        \includegraphics[width=2.245cm]{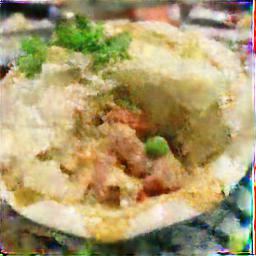}
        \centering
        \end{minipage}
        \end{subfigure}

        \vspace{0.9pt}

        \begin{subfigure}{\linewidth}
        \centering
        \begin{minipage}[t]{0.1195\linewidth}
        \centering
        \includegraphics[width=2.245cm]{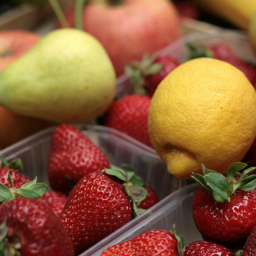}
        \caption*{\textbf{Original}}
        \centering
        \end{minipage}
        \begin{minipage}[t]{0.1195\linewidth}
        \centering
        \includegraphics[width=2.245cm]{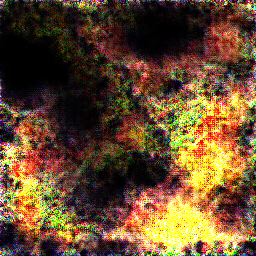}
        \caption*{\textbf{IG}}
        \centering
        \end{minipage}
        \begin{minipage}[t]{0.1195\linewidth}
        \centering
        \includegraphics[width=2.245cm]{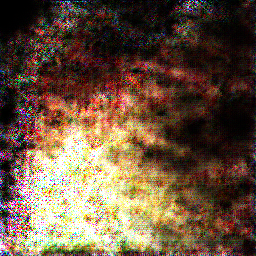}
        \caption*{\textbf{GI}}
        \centering
        \end{minipage}
        \begin{minipage}[t]{0.1195\linewidth}
        \centering
        \includegraphics[width=2.245cm]{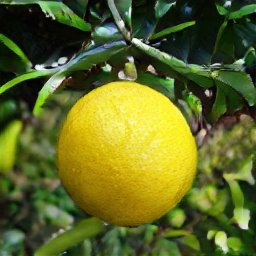}
        \caption*{\textbf{GGL}}
        \centering
        \end{minipage}
        \begin{minipage}[t]{0.1195\linewidth}
        \centering
        \includegraphics[width=2.245cm]{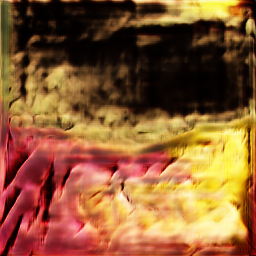}
        \caption*{\textbf{GIAS}}
        \centering
        \end{minipage}
        \begin{minipage}[t]{0.1195\linewidth}
        \centering
        \includegraphics[width=2.245cm]{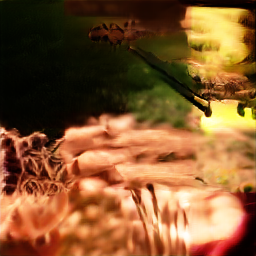}
        \caption*{\textbf{GIFD}}
        \centering
        \end{minipage}
        \begin{minipage}[t]{0.1195\linewidth}
        \centering
        \includegraphics[width=2.245cm]{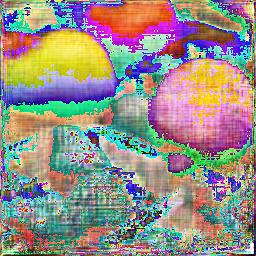}
        \caption*{\textbf{GION}}
        \centering
        \end{minipage}
        \begin{minipage}[t]{0.1195\linewidth}
        \centering
        \includegraphics[width=2.245cm]{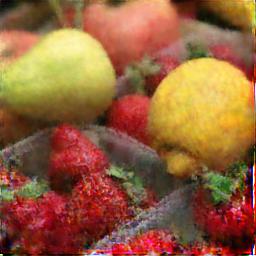}
        \caption*{\textbf{GI-NAS}}
        \centering
        \end{minipage}
        \end{subfigure}

    \caption{More qualitative results of GI-NAS and state-of-the-art gradient inversion methods on ImageNet ($256 \times 256$) with the larger batch size $B = 32$.}
    \label{visualization_imagenet_more}
\end{figure*}

We provide more visualized comparisons in Figure \ref{visualization_imagenet_more}. We observe that all the other methods struggle to conduct the reconstruction attacks, while our proposed GI-NAS again realizes the best performance in terms of visual fidelity.

\end{document}